\documentclass[10pt,twocolumn,letterpaper]{article}

\usepackage[pagenumbers]{cvpr} % To force page numbers, e.g. for an arXiv version

\usepackage[dvipsnames]{xcolor}

\newcommand{\projectcodeurl}{https://github.com/yunik1004/SAiD}
\newcommand{\projectpageurl}{https://yunik1004.github.io/SAiD}

\definecolor{cvprblue}{rgb}{0.21,0.49,0.74}
\usepackage[pagebackref,breaklinks,colorlinks,citecolor=cvprblue]{hyperref}

\usepackage{url}
\usepackage{graphicx}
\usepackage{amssymb}
\usepackage{enumitem}
\usepackage{booktabs}
\usepackage{comment}
\usepackage{bm}

\def\paperID{13160} % *** Enter the Paper ID here
\def\confName{CVPR}
\def\confYear{2024}

\title{SAiD: Speech-driven Blendshape Facial Animation with Diffusion}

\author{Inkyu Park\\
KRAFTON\\
{\tt\small inkyupark@krafton.com}
\and
Jaewoong Cho\\
KRAFTON\\
{\tt\small jwcho@krafton.com}
}

\begin{document}
\maketitle
Speech-driven 3D facial animation is challenging due to the scarcity of large-scale visual-audio datasets despite extensive research.
Most prior works, typically focused on learning regression models on a small dataset using the method of least squares, encounter difficulties generating diverse lip movements from speech and require substantial effort in refining the generated outputs.
To address these issues, we propose a speech-driven 3D facial animation with a diffusion model (SAiD), a lightweight Transformer-based U-Net with a cross-modality alignment bias between audio and visual to enhance lip synchronization.
Moreover, we introduce BlendVOCA, a benchmark dataset of pairs of speech audio and parameters of a blendshape facial model, to address the scarcity of public resources.
Our experimental results demonstrate that the proposed approach achieves comparable or superior performance in lip synchronization to baselines, ensures more diverse lip movements, and streamlines the animation editing process.

\section{Introduction}\label{sec:intro}

Speech-driven 3D facial animation has significantly enhanced human-virtual character interaction in diverse applications, including games, movies, and virtual reality platforms~\citep{meta2018oculus, edwards2020jali, taylor2017deep, adobe2018characteranimator}.
This improvement in the realism of characters emerges from tightly synchronizing speech with lip movements.
However, obtaining 3D facial animation data by motion capture is more expensive and time-consuming compared to massive 2D human face video data, consequently limiting the availability of comprehensive audio-visual datasets.
Nevertheless, there has been a proliferation of deep learning-based algorithms mapping speech audio to these 3D face meshes, a prominent three-dimensional representation of faces using a collection of vertices, edges, and faces.
There are two main approaches: one maps speech signals directly to the vertex coordinates of face meshes~\citep{cudeiro2019capture, richard2021audio, fan2022faceformer, xing2023codetalker}, 
while the other predicts coefficients associated with the face mesh, capturing essential facial deformations with fewer parameters~\citep{pham2017speech, pham2018end, peng2023emotalk}.

Most prior works typically employ regression models trained on a small-scale audio-visual dataset using the method of least squares. Despite generating plausible lip movements from speech audio, this approach still raises the following challenges: 1) not adequately capturing the inherent one-to-many relationship between speech and lip movements, and 2) requiring substantial efforts in editing the generated facial animation. If we adjust a segment of facial animation, we cannot automatically maintain the continuity of lip movement over time with the regression models. In these aspects, diffusion models can be a better candidate for speech-driven 3D facial animation. Diffusion models~\citep{sohl2015deep, ho2020denoising, dhariwal2021diffusion} have attracted considerable attention due to promising performances in generation in visual and audio domains~\citep{kong2020diffwave, jeong2021diff, liu2022diffsinger, ramesh2022hierarchical, rombach2022high, saharia2022photorealistic, podell2023sdxl} and the naturalness of image inpainting~\citep{Lugmayr2022RePaintIU}. They allow for generating new segments similar to the original segments based on specific conditions. However, the diffusion-based approach is underexplored in speech-driven 3D facial animation.

It motivates us to explore learning a diffusion model for speech-driven 3D facial animation, which can generate diverse lip-sync animation and maintain overall continuity after adjusting a segment of animation on a small-scale dataset. Specifically, we focus on the blendshape facial model~\citep{lewis2014practice}, which encapsulates facial animations via a small set of parameters and facilitates editing animation.  

We then propose \underline{S}peech-driven blendshape facial \underline{A}nimation w\underline{i}th Diffusion (SAiD), a lightweight Transformer-based UNet model crafted to generate blendshape facial model coefficients combined with a pre-trained speech encoder. 
We use the absolute error instead of the conventional squared error during the training.
The absolute error tends to reduce the perceptual distance between sampled and original data more effectively than the squared error~\citep{Saharia2021PaletteID}.
As a result, it assists in producing realistic facial animations, even when working with a limited dataset.
Next, we introduce the noise-level velocity loss that can be directly applied to the diffusion model training.
We prove that it is equivalent to the velocity loss of the sampled data.

\begin{figure*}[t]
    \centering
    \includegraphics[width=0.8\textwidth]{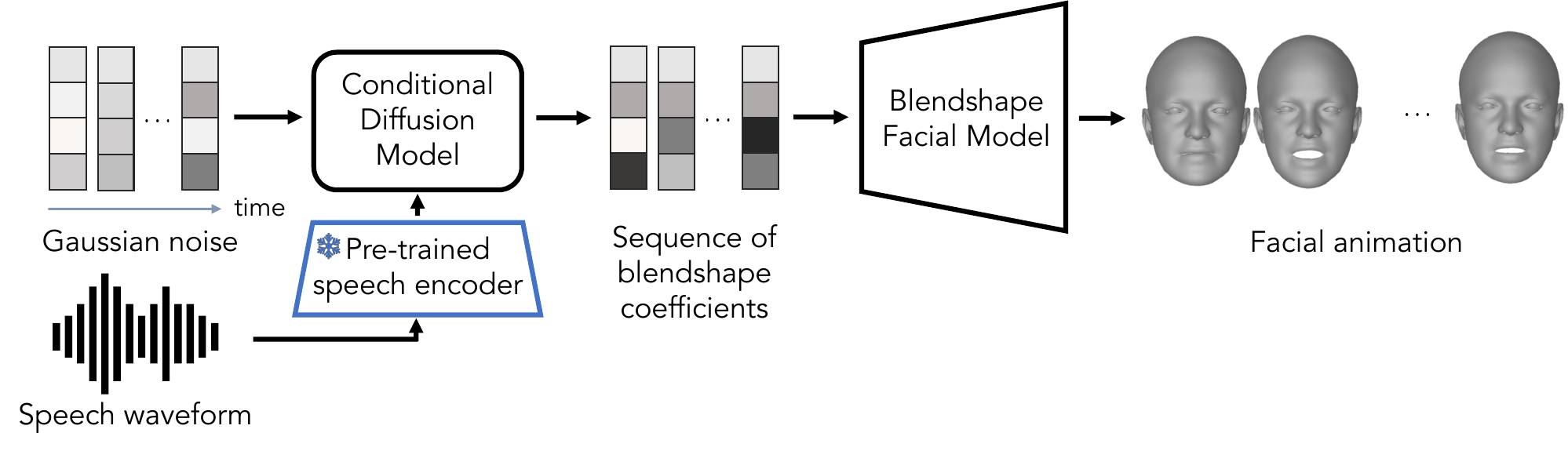}
    \caption{
        \textbf{Overview of SAiD.}
        The conditional diffusion model generates the sequence of blendshape coefficients from the Gaussian noise conditioned on the speech waveform.
        After that, generated blendshape coefficients are converted into the facial animation using the blendshape facial model.
    }
    \label{fig:overview}
\end{figure*}

To this end, we introduce the BlendVOCA dataset, a new high-quality speech-blendshape facial animation benchmark dataset built upon VOCASET~\citep{cudeiro2019capture}, a speech-facial mesh animation benchmark dataset.
It does not only address the scarcity of available datasets for exploring speech-driven blendshape-based models but also enables the direct comparison of the performance of blendshape-based and vertex-based approaches.
Compared to 3D-ETF~\citep{peng2023emotalk}, a recently released blendshape-based talking face dataset, BlendVOCA possesses the following advantages:
1) Utilization of VOCASET ensures the identical training setting with vertex-based baseline methods.
2) With its various blendshape facial models, BlendVOCA enables the diverse testing and quality assessment of blendshape coefficients.

Our extensive experiments demonstrate that SAiD outperforms baselines in terms of generalization, diversity, and smoothness of lip motions after editing.

In summary, our contributions are as follows:
\begin{itemize}
    \item \textbf{Blendshape-based benchmark dataset (BlendVOCA)}: {
        We provide a publicly accessible benchmark dataset composed of the blendshape facial model and pairs of blendshape coefficients and speech audio.
    }
    \item \textbf{Blendshape-based diffusion model (SAiD)}: {
        We propose the lightweight conditional diffusion model to solve the one-to-many speech-driven blendshape facial animation problem.
        It allows us to produce a variety of outputs and subsequently edit them utilizing the same model.
    }
    \item \textbf{Extensive experiments}: {
        We extensively evaluate our model with diverse evaluation metrics. We demonstrate the superiority of SAiD over baselines. Our code and dataset are available at \url{\projectcodeurl}.
    }
\end{itemize}

\section{Preliminaries}\label{sec:prelim}

\subsection{Mesh and Motion}

In 3D object representation, a \emph{mesh} denotes a collection of i) vertices with their positions, ii) edges connecting two vertices, and iii) faces consisting of vertices and edges. One can generate diverse 3D representations by modifying the positions of vertices from the mesh, which is called ``\emph{deformation} of the mesh''.
Furthermore, the term \emph{motion} refers to changes in the mesh over time, and one can represent motion as a sequence of vertex positions for each time step.

\subsection{Blendshapes and Blendshape Coefficients}\label{sec:prelim:blendshape}

The blendshape facial model~\citep{lewis2014practice} is a widely used linear model for representing diverse 3D face structures.
It uses a linear combination of multiple meshes which comprise 1) a template mesh and 2) \emph{blendshapes} representing deformed versions of the template mesh.
The template mesh typically illustrates a neutral facial expression, while the blendshapes depict different facial movements such as jaw opening, eye closing, and more.

Let $\bm{b}_0\in\mathbb{R}^{3M}$ denote the position vector of the template mesh, including XYZ coordinates of $M$ vertices, and $\bm{b}_1, \cdots, \bm{b}_K\in\mathbb{R}^{3M}$ indicate position vectors for the blendshapes. 
We can express a feasible representation through the blendshape coefficient as $\bm{b}_0 + \sum_{k=1}^{K} u_k (\bm{b}_k - \bm{b}_0)$.
Thus, simply altering the sequence of blendshape coefficients allows us to generate various facial motions.
In general, blendshape coefficients are deemed independent of the template mesh.
It implies that we can use identical blendshape coefficients across various facial models to represent the same facial expression.

\subsection{Deformation Transfer}

Deformation transfer~\citep{sumner2004deformation} is a computational algorithm to transfer the deformation observed in a source mesh to a different target mesh.
It requires a subset of the vertex correspondence map between the source and target meshes.

The initial phase of the process involves the creation of a face correspondence map between the two meshes based on the corresponding vertices.
After that, we can find the deformation on the target mesh by solving the optimization problem designed to maintain the transformation similarity between corresponding faces.

\subsection{Conditional Diffusion Model}\label{sec:prelim:diffusion}

Given the data distribution $q (\bm{x}_{0}, \bm{c})$, with $\bm{x}_0$ as the target data and $\bm{c}$ indicating a condition, the conditional diffusion model~\citep{sohl2015deep, dhariwal2021diffusion} is the latent variable model that approximates the conditional distribution $q(\bm{x}_{0} | \bm{c})$ using the Markov denoising process with a learned Gaussian transition starting from the random noise.
It undergoes training to approximate the trajectory of the Markov noising process, starting from the data $\bm{x}_0$, with a fixed Gaussian transition.

The sequence of denoising autoencoders, $\bm{\epsilon}_{\bm{\theta}}$, is commonly utilized to model the denoising process.
These autoencoders undergo training to predict the noise $\bm{\epsilon}$ in the input at each timestep $t$ of the denoising process.
\citet{ho2020denoising} formulates the training objective as follows:
\begin{align}\label{expr:prelim:obj}
    \mathcal{L}_{\textrm{simple}} (\bm{\theta}) = \mathbb{E}_{q, t, \bm{\epsilon}} \left[ \lVert \bm{\epsilon} - \bm{\epsilon}_{{\bm{\theta}}} (\bm{x}_t , \bm{c}, t ) \rVert^2 \right],
\end{align}
where $\bm{x}_t = \sqrt{\bar{\alpha}_t} \bm{x}_0 + \sqrt{1 - \bar{\alpha}_t} \bm{\epsilon}$ and $\bar{\alpha}_t$ is a hyperparameter related to the variance of the noising process.

\section{Related Works}\label{sec:related}

\subsection{Speech-driven 3D Facial Animation}

Speech-driven 3D facial animation is a long-standing area in computer graphics, broadly categorized into parameter-based and vertex-based methodologies.

Parameter-based animation utilizes a sequence of animation-related parameters derived from input speech.
In the past, research works~\citep{cohen2001animated, taylor2012dynamic, xu2013practical, edwards2016jali} used explicit rules to form connections between phonemes and visemes.
For instance, JALI~\citep{edwards2016jali} employs a procedural approach to animate a viseme~\citep{fisher1968confusions}-based rig with two anatomical actions: lip closure and jaw opening.
VisemeNet~\citep{zhou2018visemenet} extended JALI
by incorporating a three-stage LSTM~\citep{hochreiter1997long} for predicting JALI's parameters.
\citet{pham2017speech, pham2018end} also introduced a deep-learning-based regression model that animates head rotation and FaceWarehouse~\citep{cao2013facewarehouse}-based blendshape coefficients using audio features from the video dataset.
In a recent development, EmoTalk~\citep{peng2023emotalk} applied an emotion-disentangling technique to enhance the emotional expressions in a talking face.
Despite numerous research efforts, parameter-based animation faces a shortage of high-quality, publicly accessible data and parametric facial models capable of producing facial meshes given specific parameters.
Moreover, most parameter-based approaches cannot generate diverse outputs since they adopt regression models.

An alternative and increasingly popular approach is vertex-based animation.
The recent introduction of high-resolution 3D facial motion data~\citep{cudeiro2019capture} has propelled this field forward.
VOCA~\citep{cudeiro2019capture} was the first model capable of being applied to unseen subjects without requiring retargeting, effectively decoupling identity from facial motion.
MeshTalk~\citep{richard2021meshtalk} disentangled the facial features related/unrelated to audio using a categorical latent space to synthesize audio-uncorrelated facial features.
FaceFormer~\citep{fan2022faceformer}, on the other hand, implemented a Transformer~\citep{vaswani2017attention}-based autoregressive model to capture long-term audio context.
Recently, CodeTalker~\citep{xing2023codetalker} framed the animation generation task as a code query task in the discrete space.
It improves motion quality by reducing the uncertainty associated with cross-modal mapping.
While these studies provide a more detailed representation than parameter-based models, modifying their outputs is challenging since it requires vertex-level adjustments.
Next, they are not generalizable since they can only animate the mesh sharing the same topology as the training data.
Lastly, the models are constrained to the training identities, so they cannot generate outputs in more diverse styles.

Concurrent with our work,~\citet{stan2023facediffuser} introduced an autoregressive diffusion model-based approach.
Our work differs by 1) employing a velocity loss to reduce the jitter in lip movements, and 2) utilizing a non-autoregressive model for a faster denoising process.

\subsection{Diffusion Model for Motion Generation Tasks}

Researchers have proposed various diffusion-based methods to address challenges in motion generation tasks.
The most active research area is text-driven human motion generation, which synthesizes human movements, either joint rotations or positions, based on text input.
Pioneering studies, like \citet{tevet2023human, zhang2022motiondiffuse, kim2023flame}, have introduced diffusion-based frameworks to enhance the quality and diversity of motion generation.
Specifically, the motion diffusion model (MDM)~\citep{tevet2023human} uses a Transformer encoder for direct motion prediction, and many follow-up studies have widely adopted this model.
In particular, priorMDM~\citep{shafir2023human} applied MDM as a generative prior, and PhysDiff~\citep{yuan2022physdiff} introduced physical guidance to MDM to generate more realistic outcomes.

Beyond the scope of text-driven models, researchers have recently extended the application of diffusion models to handle audio-driven motion generation tasks.
For instance, EDGE~\citep{tseng2023edge} transforms music into corresponding dance movements, while DiffGesture~\citep{Zhu_2023_CVPR} produces upper-body gestures based on speech audio.

\section{{S}peech-driven blendshape facial {A}nimation w{i}th {D}iffusion (SAiD)}\label{sec:method}

 In this section, we introduce BlendVOCA, a new benchmark speech-blendshape dataset built upon VOCASET~\citep{cudeiro2019capture} to address the scarcity of an appropriate dataset for exploring speech-driven blendshape facial animation (\cref{sec:method:dataset}), and propose SAiD, a conditional diffusion-based generative model for generating speech-driven blendshape coefficients (\cref{sec:method:model}).  

\subsection{BlendVOCA: Speech-Blendshape Facial Animation Dataset}\label{sec:method:dataset}

\begin{figure*}[t]
    \centering
    \includegraphics[width=0.8\textwidth]{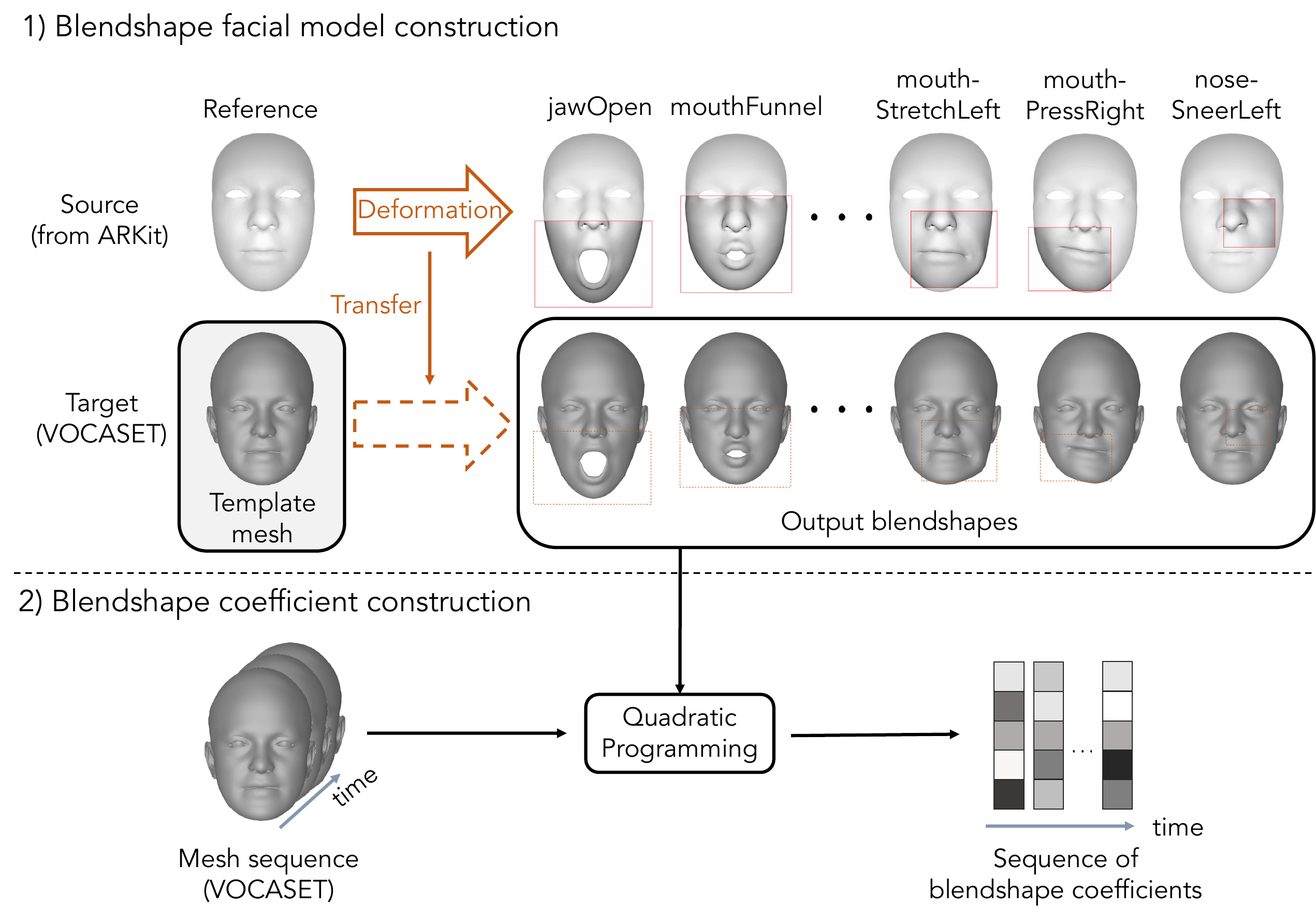}
    \caption{
        \textbf{BlendVOCA construction process.} The process unfolds in two steps: 
        1) We transfer deformations of the reference mesh from ARKit~\citep{arkit} to 12 template meshes of VOCASET~\citep{cudeiro2019capture} by applying the algorithm introduced by~\cite{sumner2004deformation}, which produce 32 output blendshape meshes for each template mesh;
        2) and then generate blendshape coefficients by solving quadratic programming problem in \cref{eq:qp}.
    }
    \label{fig:dataset_const_overview}
\end{figure*}

We construct a new dataset of template meshes and 32 blendshapes of 12 speakers, each providing approximately 40 instances of speech audios in English (each 3-5 seconds) and corresponding blendshape coefficients, all tracked over time.
To construct the dataset, we use VOCASET, which includes the template meshes of 12 speakers and speech audios, along with their synchronized facial meshes, captured at 60 frames per second for each speaker.
Our transformation of VOCASET into BlendVOCA unfolds in two steps: 1) we generate blendshapes from each template mesh by applying deformation transfer~\citep{sumner2004deformation} to the ARKit~\citep{arkit} source model, and 2) obtain blendshape coefficients by solving the optimization problem.
\cref{fig:dataset_const_overview} illustrates the overall dataset construction process.

\subsubsection{Blendshape Facial Model Construction}\label{sec:method:dataset:bfm}

Recall that deformation transfer is an algorithm used to transfer the deformation of the source blendshape model onto the target mesh. 
We use ARKit as our source model to create the blendshape facial models.
ARKit provides 52 blendshapes~\citep{arkitblendshape} based on ARSCNFaceGeometry~\citep{ARSCNFaceGeometry}, and we select 32 blendshapes that represent crucial facial features during the speech, such as those corresponding to the mouth, jaw, cheeks, nose, and so on.

Before applying deformation transfer, we preprocess the target template meshes from VOCASET by eliminating elements such as the eyeballs, ears, back of the head, and neck regions. This removal enhances stability during the blendshape facial model generation process. We reattach the removed parts to the generated blendshape meshes except for the neck.
We also construct a subset of the vertex correspondence map between the source and target meshes.
We select 68 vertices from each mesh corresponding to facial landmarks defined by~\citet{sagonas2013300}.
After that, we run the deformation transfer algorithm to get the deformation in the target mesh.
By applying the deformation, we construct the blendshape of the target mesh.
We provide the rendered images of constructed blendshape meshes in the supplementary material (\cref{fig:app:blendshape}).

\subsubsection{Blendshape Coefficient Construction}\label{sec:method:dataset:qp}

Next, we extract blendshape coefficients using both constructed blendshapes and mesh sequences in VOCASET.
To this end, we formulate a quadratic programming (QP) problem to derive the sequence of blendshape coefficients from the mesh representations of facial motion in VOCASET.
Let $\bm{p}^{1:N} = (\bm{p}^{1}, \bm{p}^{2}, \cdots, \bm{p}^{N})$ be the sequence of the position vectors corresponding to a facial motion of $N$ discrete timesteps (also known as the frame).
The goal is to obtain the smooth sequence of blendshape coefficients $\bm{u}^{1:N} = (\bm{u}^{1}, \bm{u}^{2}, \cdots, \bm{u}^{N})$ that can approximate $\bm{p}^{1:T}$ via the blendshape facial model constructed in \cref{sec:method:dataset:bfm}, where $\bm{u}^{n} = [u_1^n, u_2^n, \cdots, u_K^n]^{\intercal}$ denotes the vector, including every blendshape coefficient at the $n$-th frame.
We impose the smoothness of the blendshape coefficients over time by limiting the maximum change between two adjacent timesteps.
Therefore, we express the optimization problem as follows:
\begin{align}\label{eq:qp}
\begin{split}
\min_{\bm{u}^{1:N}} \quad & \sum_{n=1}^{N} \lVert \bm{p}^{n} - (\bm{b}_0 + \sum_{k=1}^{K} u_{k}^{n} (\bm{b}_k - \bm{b}_0)) \rVert_{2}^{2}\\
\textrm{s.t.} \quad & 0 \le u^{n}_{k} \le 1,
\quad | u^{n+1}_k - u^{n}_k | \le \delta,
\end{split}
\end{align}
where the hyperparameter $\delta > 0$ represents the maximum change allowed between two adjacent timesteps.
As demonstrated in the supplementary material (\cref{app:qp}), it is equivalent to the QP problem and has a unique solution.
Therefore, we can get unique optimization results using QP solvers.
We use CVXOPT~\citep{andersen2013cvxopt} as a QP solver with $\delta=0.1$.

\subsection{Conditional Diffusion Model for Generating Speech-Driven Coefficients}\label{sec:method:model}

\begin{figure*}[t]
    \centering
    \includegraphics[width=0.8\textwidth]{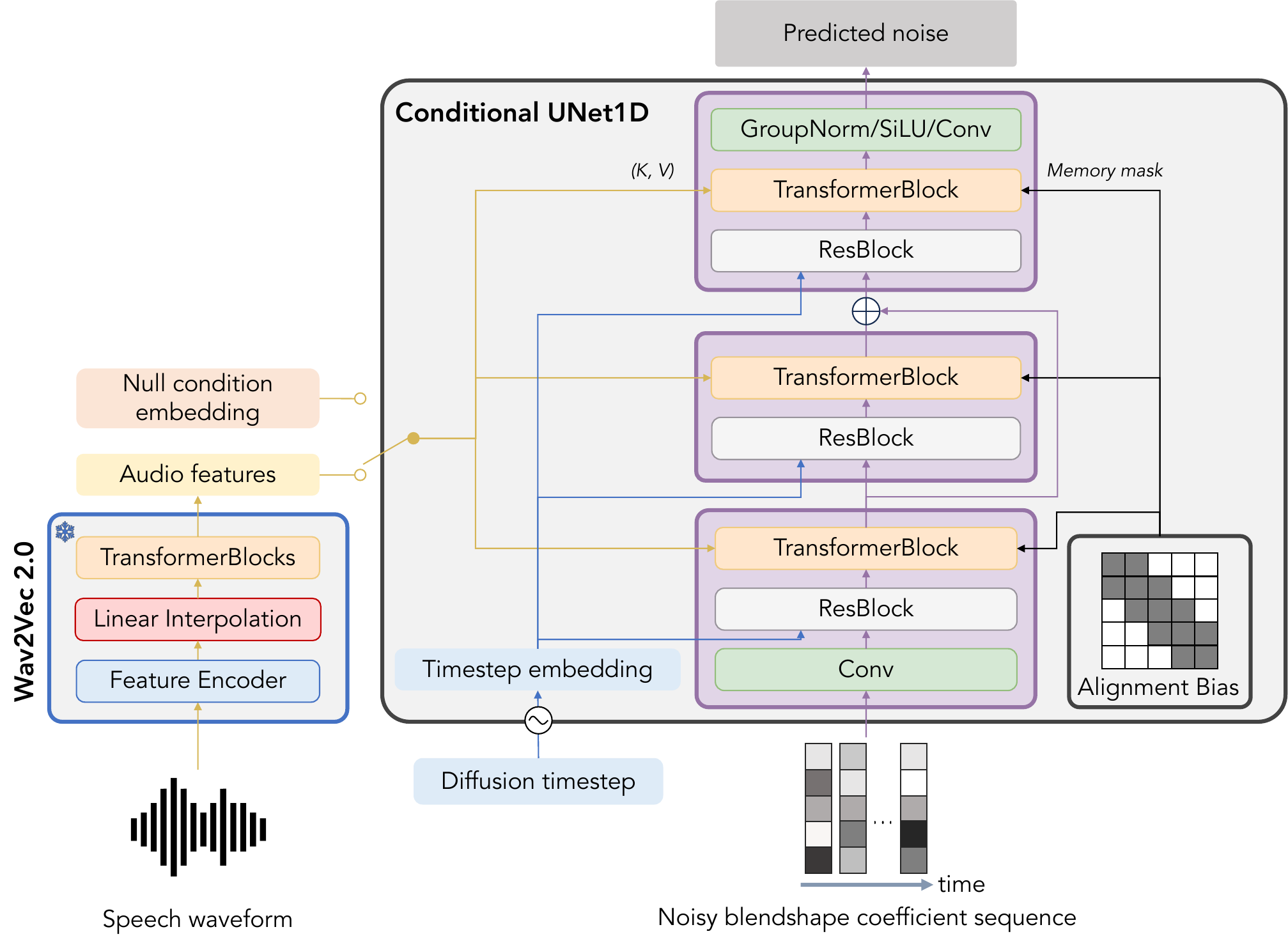}
    \caption{
        \textbf{The model architecture of SAiD.}
        SAiD predicts the noise injected into the input noisy blendshape coefficient sequence, conditioned on the speech waveform, for each diffusion timestep.
        The denoiser model is a simplified conditional UNet1D model, composed of 1 encoder block/1 middle block/1 decoder block without the downsampling and upsampling layers.
        Diffusion timestep is converted into the sinusoidal embedding and then becomes the input of each residual block in the denoiser.
        Speech waveform is converted into the audio feature vectors using the frozen pre-trained Wav2Vec 2.0 and becomes the key and value matrices of the cross-attention layer in the denoiser.
        We employ the alignment bias as a memory mask for the cross-attention layer to enhance the alignment between the speech and blendshape coefficient sequence.
        We also adopt the trainable null condition embedding for implementing the classifier-free guidance (or for the unconditional generation), providing an alternative to using the audio features.
    }
    \label{fig:said}
\end{figure*}

Given the data distribution $q (\bm{u}^{1:N}, \bm{w})$ where $\bm{w}$ denotes a speech waveform and $\bm{u}^{1:N}$ represents the corresponding sequence of the blendshape coefficients with $N$ frames, we employ the conditional diffusion model to approximate the conditional distribution $q (\bm{u}^{1:N} | \bm{w})$.

\subsubsection{Model Architecture}

Our proposed model, SAiD, in \cref{fig:said} consists of the conditional denoising UNet architecture from~\citet{rombach2022high} as a denoising autoencoder $\epsilon_{\bm{\theta}}$ and pre-trained Wav2Vec 2.0~\citep{baevski2020wav2vec} as a speech audio encoder $s$.
We modify the UNet architecture to make the input dimension 2D to 1D.
Next, 1) we remove the down/up-sampling blocks in UNet to reduce the model size, and 2) add the linear interpolation layer~\citep{fan2022faceformer} in Wav2Vec 2.0 to equalize the number of frames of the blendshape coefficient sequence and the length of the audio features.
After that, we use the audio features as keys and values of the cross-attention layers in UNet.

To synchronize the audio and the blendshape coefficient sequence, we use the alignment bias~\citep{fan2022faceformer} as a memory mask for the cross-attention layer, i.e., the additive mask for the audio encoder output.
Focusing on the audio corresponding to adjacent frames enhances the model's learning of local feature information more effectively.
Therefore, we modify the cross-attention output of the Transformer decoder block in UNet as follows:
\begin{align}\label{expr:method:cross_attn}
    \mathrm{Attention} (\bm{Q}, \bm{K}, \bm{V}) = \mathrm{softmax} (\frac{\bm{Q} \bm{K}^{\intercal}}{\sqrt{d}} + \bm{B}^{A})\bm{V},
\end{align}
where $\bm{Q}, \bm{K}, \bm{V} \in \mathbb{R}^{N \times d}$ are the query, key, and value matrices, and $\bm{B}^{A} \in \mathbb{R}^{N \times N}$ is the alignment bias:
\begin{align}\label{expr:method:bias}
    \bm{B}^{A} (i, j) =
    \begin{cases}
        0, &|i - j| \le 1\\
        -\infty, &\textrm{otherwise}
    \end{cases}.
\end{align}
\vspace{-2mm}
\subsubsection{Training}

To train the UNet model, we use the training objective similar to \cref{expr:prelim:obj}.
Instead of using the squared error, we minimize the absolute error~\citep{chen2020wavegrad, Saharia2021PaletteID, saharia2022image} between the noise and the predicted noise at each diffusion timestep.
It reduces the perceptual distance between the sample and real-data distribution compared to the squared error~\citep{Saharia2021PaletteID}, helping to produce realistic lip movements even with less data.
Therefore, the simple training objective is as follows:
\begin{align}\label{expr:method:simple}
\begin{split}
    \mathcal{L}_{\mathrm{simple}} (\bm{\theta}) = \mathbb{E}_{q, t, \bm{\epsilon}} \Bigl[\lVert \bm{\epsilon} - \epsilon_{\bm{\theta}} ( \bm{u}^{1:N}_t , s(\bm{w}), t)  \rVert_1 \Bigr],
\end{split}
\end{align}
where $\bm{\theta}$ is a trainable parameter of the UNet model, and $\bm{u}^{1:N}_t = \sqrt{\bar{\alpha}_t} \bm{u}^{1:N} + \sqrt{1 - \bar{\alpha}_t} \bm{\epsilon}$.

We apply the additional loss to reduce the jitter of the output, achieving this by minimizing the gap between the temporal difference, known as velocity, of $\bm{u}^{1:N}$ and the velocity of the denoised observation $\hat{\bm{u}}^{1:N} = (\bm{u}^{1:N}_t - \sqrt{1 - \bar{\alpha}}_t \hat{\bm{\epsilon}})  / \sqrt{\bar{\alpha}_t}$~\citep{ho2020denoising} at each diffusion timestep, where $\hat{\bm{\epsilon}}_t = \epsilon_{\bm{\theta}} (\bm{u}^{1:N}_t, s(\bm{w}), t)$:
{\small
\begin{align}\label{expr:method:tc:original}
    \mathcal{L}_{\mathrm{vel}}' (\bm{\theta}) = \mathbb{E}_{q, t, \bm{\epsilon}} \Bigl[ \sum_{n=1}^{N-1} \lVert (\bm{u}^{n+1} - \bm{u}^{n}) - (\hat{\bm{u}}^{n+1} - \hat{\bm{u}}^{n}) \rVert_1 \Bigr]&\nonumber\\
    = \mathbb{E}_{q, t, \bm{\epsilon}} \Bigl[ \sqrt{\frac{1 - \bar{\alpha}_t}{\bar{\alpha}_t}} \sum_{n=1}^{N-1} \lVert (\bm{\epsilon}^{n+1} - \bm{\epsilon}^{n}) - (\hat{\bm{\epsilon}}^{n+1}_t - \hat{\bm{\epsilon}}^{n}_t) \rVert_1 \Bigr]&,
\end{align}
}
where $\bm{\epsilon}^{n}$ and $\hat{\bm{\epsilon}}^{n}_t$ denote the $n$-th frame component of $\bm{\epsilon}$ and $\hat{\bm{\epsilon}}_t$.
We use the reweighted version of \cref{expr:method:tc:original} by removing the coefficients of each term, which is equivalent to the noise-level velocity loss:
{\small
\begin{align}\label{expr:method:tc:reweight}
    \mathcal{L}_{\mathrm{vel}} (\bm{\theta}) = \mathbb{E}_{q, t, \bm{\epsilon}} \Bigl[ \sum_{n=1}^{N-1} \lVert (\bm{\epsilon}^{n+1} - \bm{\epsilon}^{n}) - (\hat{\bm{\epsilon}}^{n+1}_t - \hat{\bm{\epsilon}}^{n}_t) \rVert_1 \Bigr].
\end{align}}

As a result, our training loss is:
\begin{align}\label{expr:method:loss_full}
    \mathcal{L} (\bm{\theta}) = \mathcal{L}_{\mathrm{simple}} (\bm{\theta}) + \mathcal{L}_{\mathrm{vel}} (\bm{\theta}).
\end{align}
\subsubsection{Conditional Sampling}

We can apply various sampling methods, such as DDPM~\citep{ho2020denoising} or DDIM~\citep{Song2020DenoisingDI}, with classifier-free guidance~\citep{ho2022classifier} to execute conditional sampling.
For each sampling step $t$, we replace the predicted noise with the following linear combination of the conditional and unconditional estimates:
\begin{align}
\begin{split}
    \Tilde{\epsilon}_\theta ( \bm{u}^{1:N}_t ,&~ s(\bm{w}),t) = \epsilon_\theta ( \bm{u}^{1:N}_t , s(\bm{w}), t) \\
    &+ \gamma ( \epsilon_\theta ( \bm{u}^{1:N}_t , s(\bm{w}), t) - \epsilon_\theta ( \bm{u}^{1:N}_t , \bm{\emptyset}, t)),
\end{split}
\end{align}
where $\bm{u}^{1:N}_t$ indicates the intermediate noisy blendshape coefficient sequence at sampling step $t$, $\bm{\emptyset}$ denotes the embedding of the null condition, and $\gamma \ge 0$ is a hyperparameter that controls the strength of the classifier-free guidance.
We train $\bm{\emptyset}$ during the training stage by randomly replacing the condition $s(\bm{w})$ with $\bm{\emptyset}$ at a $0.1$ probability.
Empirically, we set $\gamma = 2$ to get the best result.

\subsubsection{Editing of the Blendshape Coefficient Sequence}\label{sec:method:model:edit}

SAiD also functions as an editing tool for the blendshape coefficient sequence like a typical diffusion model~\citep{Lugmayr2022RePaintIU}.

Let $\bm{u}^{1:N}_{\mathrm{ref}}$ represent the blendshape coefficient sequence we aim to modify, and $\bm{w}$ the corresponding speech waveform.
Suppose we have a binary mask $\bm{m}$, which indicates the parts that need editing with $0$ and the rest with $1$.
Under these conditions, SAiD can regenerate the unmasked areas of $\bm{u}^{1:N}_{\mathrm{ref}}$ by updating $\bm{u}^{1:N}_{t}$, which represents the intermediate noisy blendshape coefficient sequence at each sampling step $t$, with the following adjustment:
\begin{align}\label{expr:method:edit}
\begin{split}
    \Tilde{\bm{u}}^{1:N}_{t} &= (\bm{1} - \bm{m}) \circ \bm{u}^{1:N}_{t}\\
    &+ \bm{m} \circ (\sqrt{\bar{\alpha}_t} \bm{u}^{1:N}_{\mathrm{ref}} + \sqrt{1 - \bar{\alpha}_t} \bm{\epsilon}),
\end{split}
\end{align}
where $\circ$ indicates the element-wise product operator.

\section{Experiments}\label{sec:experiment}

\begin{table*}[ht]
    \centering
    \begin{subtable}[h]{\textwidth}
        \centering
        \begin{tabular}{l|ccccc}
            \toprule
            \textbf{Methods} \hspace{3.7cm} & \textbf{AV Offset $\rightarrow$} & \textbf{AV Confidence $\uparrow$} & \textbf{Multimodality $\uparrow$} & \textbf{FD $\downarrow$} & \textbf{WInD $\downarrow$}\\
            \midrule
            Ground-Truth & $-1.038$ & $4.874$ & N/A & $0.000$ & $1.120^{\pm0.450}$  \\
            \midrule
            SAiD (Ours) & $\mathbf{-1.025}$ & $\mathbf{5.575}$ & $\mathbf{3.817}$ & $\mathbf{6.791}$ & $\underline{10.344^{\pm0.127}}$ \\
            \midrule
            end2end\_AU\_speech~\citep{pham2018end} & $0.785$ & $0.887$ & $0.000$ & $12.307$ & $14.070^{\pm0.076}$\\
            VOCA~\citep{cudeiro2019capture} + QP & $\underline{-0.891}$ & $3.117$ & $\underline{2.899}$ & $45.555$ & $52.403^{\pm0.402}$\\
            MeshTalk~\citep{richard2021meshtalk} + QP & $-1.532$ & $4.425$ & $0.000$ & $11.106$ & $13.161^{\pm0.046}$\\
            FaceFormer~\citep{fan2022faceformer} + QP & $-0.723$ & $\underline{5.346}$ & $2.490$ & $10.265$ & $14.102^{\pm0.080}$\\
            CodeTalker~\citep{xing2023codetalker} + QP & $-1.476$ & $5.256$ & $1.407$ & $\underline{6.862}$ & $\mathbf{9.813^{\pm0.067}}$\\
            \bottomrule
        \end{tabular}
    \end{subtable}
     \caption{
        \textbf{Evaluation results on the test data.}
        SAiD achieves the best results in AV offset/confidence, multimodality, and FD while taking second place in WInD. It highlights SAiD's ability to generate diverse outputs while closely aligning with the real-data distribution.
        $\uparrow$ implies the upper is better, $\downarrow$ implies the lower is better, and $\rightarrow$ implies the metric closer to the ground truth is better.
        \textbf{Bold} indicates the best result, \underline{underline} indicates the second-best result, and $\pm$ indicates the standard deviation. 
     }\label{tab:exp:eval:comp}
\end{table*}

\begin{table*}[ht]
    \centering
    \begin{subtable}[h]{\textwidth}
        \centering
        \begin{tabular}{l|ccccc}
            \toprule
            \textbf{Methods} \hspace{3.7cm} & \textbf{AV Offset $\rightarrow$} & \textbf{AV Confidence $\uparrow$} & \textbf{Multimodality $\uparrow$} & \textbf{FD $\downarrow$} & \textbf{WInD $\downarrow$}\\
            \midrule
            SAiD (Base) & $\mathbf{-1.025}$ & $\mathbf{5.575}$ & $\underline{3.817}$ & $\underline{6.791}$ & $\underline{10.344^{\pm0.127}}$\\
            \hspace{3mm} train w/ squared error & $-1.012$ & $5.279$ & $3.772$ & $\mathbf{6.601}$ & $\mathbf{9.653^{\pm0.064}}$\\
            \hspace{3mm} train w/o velocity loss & $\underline{-1.021}$ & $5.476$ & $3.736$ & $7.279$ & $10.995^{\pm0.089}$\\
            \hspace{3mm} train w/o alignment bias & $0.157$ & $1.646$ & $\mathbf{6.951}$ & $64.060$ & $69.470^{\pm0.183}$\\
            \hspace{3mm} finetune pre-trained Wav2Vec 2.0 & $-1.015$ & $\underline{5.498}$ & $2.736$ & $10.177$ & $14.723^{\pm0.174}$\\
            \bottomrule
        \end{tabular}
     \end{subtable}
     \caption{
        \textbf{Ablation study results.}
        We explore SAiD's performance variations with different architecture components and training losses.
        Our design choice achieves the top performance in AV offset/confidence and the second-best results in multimodality, FD, and WInD.
     }\label{tab:exp:eval:ablation}
\end{table*}

\subsection{Training Details of SAiD}\label{sec:experiment:training}

We employ the BlendVOCA dataset constructed using the procedure described in \cref{sec:method:dataset}. We adopt the same training/validation/test splits, specifically 8/2/2 speakers split, used by VOCA~\citep{cudeiro2019capture}.

For each training step, we randomly choose a minibatch of size $8$.
Each data in the minibatch consists of the randomly sliced blendshape coefficient sequence and a corresponding speech waveform.
To use the classifier-free guidance, we randomly replace the output audio features of the speech audio encoder into the null condition embedding with a probability of $0.1$.
We adopt diverse data augmentation strategies to amplify our training dataset.
We execute the audio augmentation by shifting the speech waveform within 1/60 second with a probability of $0.5$.
Subsequently, we swap the blendshape coefficients between the symmetric blendshapes with a probability of $0.5$.

We conduct the training on a single NVIDIA A100 (40GB) for 50,000 epochs using the AdamW~\citep{loshchilov2017decoupled} optimizer with $\beta_1=0.9$, $\beta_2 = 0.999$, a learning rate of $10^{-5}$ with a warmup, and a weight decay of $10^{-2}$.
We use EMA~\citep{tarvainen2017mean} with a decay value of $0.9999$ to update the model weights.
We select the model that minimizes the validation loss.

\subsection{Baseline Methods}

We compare SAiD with end2end\_AU\_speech~\citep{pham2018end}, which regresses the blendshape coefficients given the audio spectrogram.
Considering the limited available models for speech-driven blendshape facial animation, we further compare SAiD with the models that directly produce the mesh sequence of the facial motion.
We choose VOCA~\citep{cudeiro2019capture}, MeshTalk~\citep{richard2021meshtalk}, FaceFormer~\citep{fan2022faceformer}, and CodeTalker~\citep{xing2023codetalker} as our baseline methods for comparison.
We optimize the blendshape coefficient sequence from the generated mesh sequence by solving the QP problem described in \cref{sec:method:dataset:qp}.
See the supplementary material (\cref{app:baseline}) to check the details.

\subsection{Evaluation Metrics}

We employ five metrics to evaluate the performances of SAiD and baselines: audio-visual offset/confidence~\citep{chung2017out}, multimodality~\citep{guo2020action2motion}, Frechet distance (FD)~\citep{dowson1982frechet, heusel2017gans}, and Wasserstein inception distance (WInD)~\citep{dimitrakopoulos2020wind}.
These metrics measure 1) the synchronization between lip movements and speech (AV offset/confidence), 2) the variety in lip movements (multimodality), and 3) the similarity between the ground truth and generated samples (FD, WInD).
Details of these metrics are provided in the supplementary material (\cref{app:eval_metric,app:feature_extractor,app:exp_setting}).

\subsection{Evaluation Results}

\cref{tab:exp:eval:comp} indicates that our proposed method, SAiD, achieves the best results in AV offset/confidence, multimodality, and FD while taking second place in WInD.
It highlights SAiD's ability to generate diverse outputs while closely aligning with the real-data distribution.
We provide the qualitative demo examples at \url{\projectpageurl}.

\subsection{Facial Motion Editing}

We conduct two experiments focused on regenerating unmasked areas of the blendshape coefficient sequence while fixing the masked regions, using the method in \cref{sec:method:model:edit}.
\vspace{-4mm}
\paragraph{Motion in-betweening:} {
We mask the beginning and end blendshape coefficients and generate the intermediate values.
}
\vspace{-4mm}
\paragraph{Motion generation with blendshape-specific constraints:} {
We mask the coefficients of specific blendshapes and generate the remaining blendshape coefficients.
}

\begin{figure}
    \begin{subfigure}{\linewidth}
        \centering
        \includegraphics[width=0.95\textwidth]{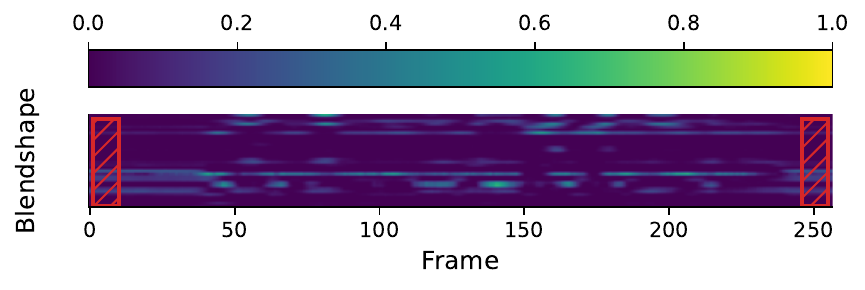}
        \caption{Motion in-betweening}
    \end{subfigure}
    \begin{subfigure}{\linewidth}
        \centering
        \includegraphics[width=0.95\textwidth]{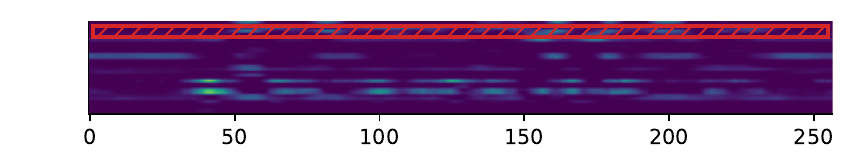}
        \caption{Motion generation with blendshape-specific constraints}
    \end{subfigure}
    \caption{
        \textbf{Motion editing.}
        Hatched boxes indicate the masked areas that should be invariant during the editing.
        SAiD can generate motions on the unmasked area using motion editing in \cref{sec:method:model:edit}.
        We provide the videos results of these editing tasks at \url{\projectpageurl}.
    }
    \label{fig:motion_edit}
\end{figure}

SAiD seamlessly generates blendshape coefficients in unmasked areas that integrate with the masked sections, as illustrated in \cref{fig:motion_edit}.
It demonstrates the editability of SAiD by leveraging the strengths of the diffusion model.

\section{Ablation Studies}

\begin{figure}[t]
    \begin{subfigure}{\linewidth}
        \centering
        \includegraphics[width=0.9\textwidth]{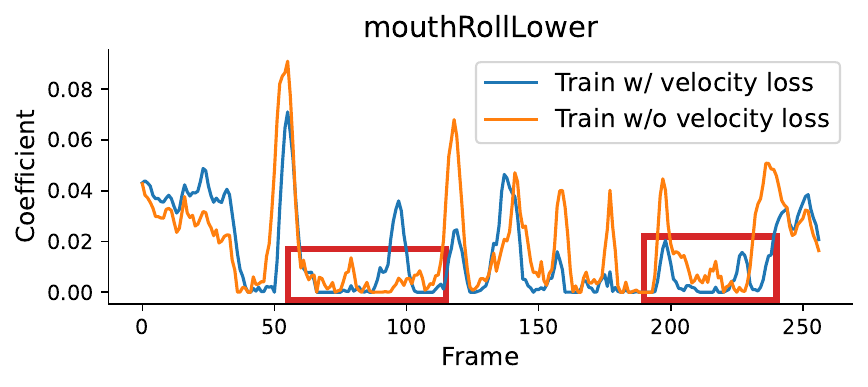}
    \end{subfigure}
    \begin{subfigure}{\linewidth}
        \centering
        \includegraphics[width=0.9\textwidth]{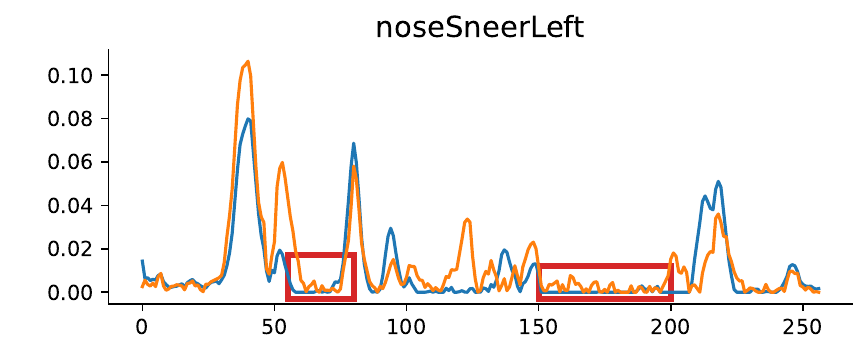}
    \end{subfigure}
    \caption{
        \textbf{Effect of the velocity loss.}
        Blue lines indicate SAiD's inference results with velocity loss training, while orange lines display results without velocity loss.
        As highlighted in the red box, the blue lines demonstrate notably reduced jitter compared to the orange lines.
    }
    \label{fig:tc:velocity}
\end{figure}

\begin{figure}[t]
    \begin{subfigure}{\linewidth}
        \centering
        \includegraphics[width=0.48\textwidth]{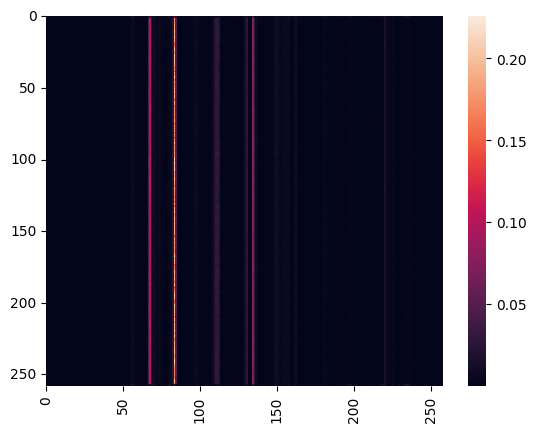}
        \includegraphics[width=0.48\textwidth]{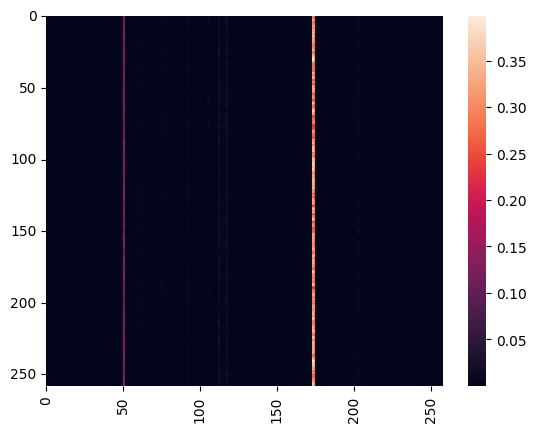}
        \caption{Cross-attention map without alignment bias}
        \label{fig:bias:without_bias}
    \end{subfigure}
    \begin{subfigure}{\linewidth}
        \centering
        \includegraphics[width=0.48\textwidth]{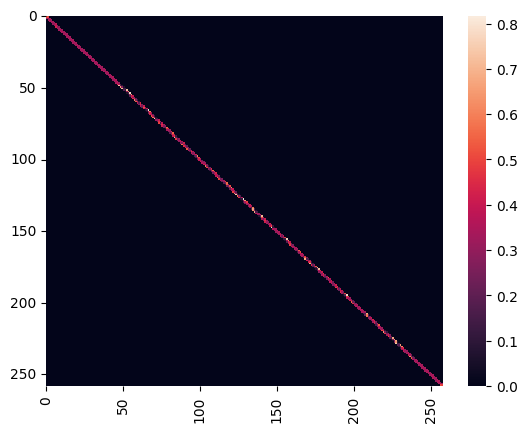}
        \includegraphics[width=0.48\textwidth]{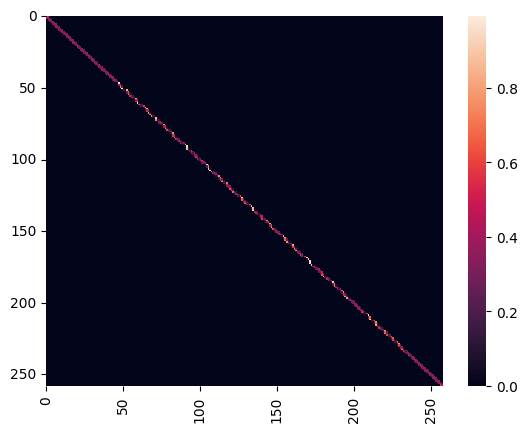}
        \caption{Cross-attention map with alignment bias}
        \label{fig:bias:with_bias}
    \end{subfigure}
    \caption{
        \textbf{Effect of the alignment bias on the cross-attention maps.}
         (a) shows that SAiD trained without alignment bias cannot learn the alignment between the audio and the blendshape coefficient sequence. (b) presents that the bias enforces the alignment between them.
    }
    \label{fig:bias}
\end{figure}

We investigate the performance of SAiD by changing architectural components and the training loss.
Evaluation results are in \cref{tab:exp:eval:ablation}.
    \vspace{-4mm}
    \paragraph{Effect of absolute error:} {
        SAiD trained with squared error reveals improved FD and WInD.
        On the other hand, it shows diminished AV offset/confidence and multimodality, aspects related to perceptual correctness.
    }
    \vspace{-4mm}
    \paragraph{Effect of noise-level velocity loss (\cref{expr:method:tc:reweight}):} {
        \cref{fig:tc:velocity} illustrates that SAiD without the velocity loss produces results with high-frequency variations, commonly known as jitter.
        It yields the decline of the overall evaluation scores.
        From this observation, we conclude that the noise-level velocity loss reduces the jitter in the inference results.
    }
    \vspace{-4mm}
    \paragraph{Effect of alignment bias (\cref{expr:method:bias}):} {
        We investigate the role of the alignment bias by comparing the cross-attention map (i.e., the softmax output of \cref{expr:method:cross_attn}) of SAiD.
        As depicted in \cref{fig:bias:without_bias}, the cross-attention map trained without the alignment bias lacks the alignment between the audio and the blendshape coefficient sequence.
        However, when trained with alignment bias, the cross-attention map clearly illustrates an alignment between these elements, as evident in \cref{fig:bias:with_bias}.
        The evaluation metrics are also significantly improved after using the alignment bias.
        Hence, the alignment bias is crucial in achieving accurate alignment.
    }
    \vspace{-4mm}
    \paragraph{Effect of speech encoder freezing:} {
        Finetuning the pre-trained Wav2Vec 2.0 can decrease the overall performance of SAiD.
        Due to the limited data in VOCASET, the encoder seems to struggle to encode general voice information and overfit easily.
    }

\section{Conclusion}\label{sec:conclusion}

We propose SAiD, a diffusion-based approach for the speech-driven 3D facial animation problem, using a lightweight blendshape-based diffusion model.
We introduce BlendVOCA, a benchmark dataset that pairs speech audio with blendshape coefficients for training a blendshape-based model.
Our experiments demonstrate that SAiD generates diverse lip movements while outperforming the existing methods in synchronizing lip movements with speech.
Moreover, SAiD showcases its capability to streamline the motion editing process.

Nevertheless, SAiD encounters a limitation: it relies on local attention in the cross-attention layer, which complicates the utilization of global information during the diffusion process.
Future work may explore aligning audio with blendshape coefficients without employing harsh alignment bias.
We might schedule the bias bandwidth across the diffusion timestep, enabling a transition from global attention initially to local attention towards the end.

{
    \small
    \bibliographystyle{ieeenat_fullname}
    \bibliography{reference}

\begin{thebibliography}{64}
\providecommand{\natexlab}[1]{#1}
\providecommand{\url}[1]{\texttt{#1}}
\expandafter\ifx\csname urlstyle\endcsname\relax
  \providecommand{\doi}[1]{doi: #1}\else
  \providecommand{\doi}{doi: \begingroup \urlstyle{rm}\Url}\fi

\bibitem[Adobe(2020)]{adobe2018characteranimator}
Adobe.
\newblock Animated lip-syncing powered by adobe ai.
\newblock 2020.

\bibitem[Andersen et~al.(2013)Andersen, Dahl, Vandenberghe,
  et~al.]{andersen2013cvxopt}
Martin~S Andersen, Joachim Dahl, Lieven Vandenberghe, et~al.
\newblock {CVXOPT}: A python package for convex optimization.
\newblock 2013.

\bibitem[Apple(2017{\natexlab{a}})]{ARSCNFaceGeometry}
Apple.
\newblock Apple developer documentation - {ARSCNFaceGeometry}.
\newblock 2017{\natexlab{a}}.

\bibitem[Apple(2017{\natexlab{b}})]{arkit}
Apple.
\newblock Apple developer documentation - {ARKit}.
\newblock 2017{\natexlab{b}}.

\bibitem[Apple(2017{\natexlab{c}})]{arkitblendshape}
Apple.
\newblock Apple developer documentation - {ARFaceAnchor.BlendShapeLocation}.
\newblock 2017{\natexlab{c}}.

\bibitem[Baevski et~al.(2020)Baevski, Zhou, Mohamed, and
  Auli]{baevski2020wav2vec}
Alexei Baevski, Yuhao Zhou, Abdelrahman Mohamed, and Michael Auli.
\newblock wav2vec 2.0: A framework for self-supervised learning of speech
  representations.
\newblock \emph{Advances in neural information processing systems},
  33:\penalty0 12449--12460, 2020.

\bibitem[Cao et~al.(2013)Cao, Weng, Zhou, Tong, and Zhou]{cao2013facewarehouse}
Chen Cao, Yanlin Weng, Shun Zhou, Yiying Tong, and Kun Zhou.
\newblock Facewarehouse: A 3d facial expression database for visual computing.
\newblock \emph{IEEE Transactions on Visualization and Computer Graphics},
  20\penalty0 (3):\penalty0 413--425, 2013.

\bibitem[Chen et~al.(2020)Chen, Zhang, Zen, Weiss, Norouzi, and
  Chan]{chen2020wavegrad}
Nanxin Chen, Yu Zhang, Heiga Zen, Ron~J Weiss, Mohammad Norouzi, and William
  Chan.
\newblock Wavegrad: Estimating gradients for waveform generation.
\newblock \emph{arXiv preprint arXiv:2009.00713}, 2020.

\bibitem[Chung and Zisserman(2017)]{chung2017out}
Joon~Son Chung and Andrew Zisserman.
\newblock Out of time: automated lip sync in the wild.
\newblock In \emph{Computer Vision--ACCV 2016 Workshops: ACCV 2016
  International Workshops, Taipei, Taiwan, November 20-24, 2016, Revised
  Selected Papers, Part II 13}, pages 251--263. Springer, 2017.

\bibitem[Cohen et~al.(2001)Cohen, Clark, and Massaro]{cohen2001animated}
Michael~M Cohen, Rashid Clark, and Dominic~W Massaro.
\newblock Animated speech: Research progress and applications.
\newblock In \emph{AVSP 2001-International Conference on Auditory-Visual Speech
  Processing}, 2001.

\bibitem[Cudeiro et~al.(2019)Cudeiro, Bolkart, Laidlaw, Ranjan, and
  Black]{cudeiro2019capture}
Daniel Cudeiro, Timo Bolkart, Cassidy Laidlaw, Anurag Ranjan, and Michael~J
  Black.
\newblock Capture, learning, and synthesis of 3d speaking styles.
\newblock In \emph{Proceedings of the IEEE/CVF Conference on Computer Vision
  and Pattern Recognition}, pages 10101--10111, 2019.

\bibitem[Dhariwal and Nichol(2021)]{dhariwal2021diffusion}
Prafulla Dhariwal and Alexander Nichol.
\newblock Diffusion models beat gans on image synthesis.
\newblock \emph{Advances in Neural Information Processing Systems},
  34:\penalty0 8780--8794, 2021.

\bibitem[Dimitrakopoulos et~al.(2020)Dimitrakopoulos, Sfikas, and
  Nikou]{dimitrakopoulos2020wind}
Panagiotis Dimitrakopoulos, Giorgos Sfikas, and Christophoros Nikou.
\newblock Wind: Wasserstein inception distance for evaluating generative
  adversarial network performance.
\newblock In \emph{ICASSP 2020-2020 IEEE International Conference on Acoustics,
  Speech and Signal Processing (ICASSP)}, pages 3182--3186. IEEE, 2020.

\bibitem[Dowson and Landau(1982)]{dowson1982frechet}
DC Dowson and BV666017 Landau.
\newblock The fr{\'e}chet distance between multivariate normal distributions.
\newblock \emph{Journal of multivariate analysis}, 12\penalty0 (3):\penalty0
  450--455, 1982.

\bibitem[Edwards et~al.(2016)Edwards, Landreth, Fiume, and
  Singh]{edwards2016jali}
Pif Edwards, Chris Landreth, Eugene Fiume, and Karan Singh.
\newblock Jali: an animator-centric viseme model for expressive lip
  synchronization.
\newblock \emph{ACM Transactions on graphics (TOG)}, 35\penalty0 (4):\penalty0
  1--11, 2016.

\bibitem[Edwards et~al.(2020)Edwards, Landreth, Pop\l{}awski, Malinowski,
  Watling, Fiume, and Singh]{edwards2020jali}
Pif Edwards, Chris Landreth, Mateusz Pop\l{}awski, Robert Malinowski, Sarah
  Watling, Eugene Fiume, and Karan Singh.
\newblock Jali-driven expressive facial animation and multilingual speech in
  cyberpunk 2077.
\newblock In \emph{ACM SIGGRAPH 2020 Talks}, New York, NY, USA, 2020.
  Association for Computing Machinery.

\bibitem[Fan et~al.(2022)Fan, Lin, Saito, Wang, and Komura]{fan2022faceformer}
Yingruo Fan, Zhaojiang Lin, Jun Saito, Wenping Wang, and Taku Komura.
\newblock Faceformer: Speech-driven 3d facial animation with transformers.
\newblock In \emph{Proceedings of the IEEE/CVF Conference on Computer Vision
  and Pattern Recognition}, pages 18770--18780, 2022.

\bibitem[Fisher(1968)]{fisher1968confusions}
Cletus~G Fisher.
\newblock Confusions among visually perceived consonants.
\newblock \emph{Journal of speech and hearing research}, 11\penalty0
  (4):\penalty0 796--804, 1968.

\bibitem[Fu et~al.(2019)Fu, Li, Liu, Gao, Celikyilmaz, and
  Carin]{Fu2019CyclicalAS}
Hao Fu, Chunyuan Li, Xiaodong Liu, Jianfeng Gao, Asli Celikyilmaz, and Lawrence
  Carin.
\newblock Cyclical annealing schedule: A simple approach to mitigating kl
  vanishing.
\newblock In \emph{North American Chapter of the Association for Computational
  Linguistics}, 2019.

\bibitem[Guo et~al.(2020)Guo, Zuo, Wang, Zou, Sun, Deng, Gong, and
  Cheng]{guo2020action2motion}
Chuan Guo, Xinxin Zuo, Sen Wang, Shihao Zou, Qingyao Sun, Annan Deng, Minglun
  Gong, and Li Cheng.
\newblock Action2motion: Conditioned generation of 3d human motions.
\newblock In \emph{Proceedings of the 28th ACM International Conference on
  Multimedia}, pages 2021--2029, 2020.

\bibitem[Heusel et~al.(2017)Heusel, Ramsauer, Unterthiner, Nessler, and
  Hochreiter]{heusel2017gans}
Martin Heusel, Hubert Ramsauer, Thomas Unterthiner, Bernhard Nessler, and Sepp
  Hochreiter.
\newblock Gans trained by a two time-scale update rule converge to a local nash
  equilibrium.
\newblock \emph{Advances in neural information processing systems}, 30, 2017.

\bibitem[Ho and Salimans(2022)]{ho2022classifier}
Jonathan Ho and Tim Salimans.
\newblock Classifier-free diffusion guidance.
\newblock \emph{arXiv preprint arXiv:2207.12598}, 2022.

\bibitem[Ho et~al.(2020)Ho, Jain, and Abbeel]{ho2020denoising}
Jonathan Ho, Ajay Jain, and Pieter Abbeel.
\newblock Denoising diffusion probabilistic models.
\newblock \emph{Advances in Neural Information Processing Systems},
  33:\penalty0 6840--6851, 2020.

\bibitem[Hochreiter and Schmidhuber(1997)]{hochreiter1997long}
Sepp Hochreiter and J{\"u}rgen Schmidhuber.
\newblock Long short-term memory.
\newblock \emph{Neural computation}, 9\penalty0 (8):\penalty0 1735--1780, 1997.

\bibitem[Jeong et~al.(2021)Jeong, Kim, Cheon, Choi, and Kim]{jeong2021diff}
Myeonghun Jeong, Hyeongju Kim, Sung~Jun Cheon, Byoung~Jin Choi, and Nam~Soo
  Kim.
\newblock Diff-tts: A denoising diffusion model for text-to-speech.
\newblock \emph{arXiv preprint arXiv:2104.01409}, 2021.

\bibitem[Kim et~al.(2023)Kim, Kim, and Choi]{kim2023flame}
Jihoon Kim, Jiseob Kim, and Sungjoon Choi.
\newblock Flame: Free-form language-based motion synthesis \& editing.
\newblock In \emph{Proceedings of the AAAI Conference on Artificial
  Intelligence}, pages 8255--8263, 2023.

\bibitem[Kingma and Welling(2013)]{kingma2013auto}
Diederik~P Kingma and Max Welling.
\newblock Auto-encoding variational bayes.
\newblock \emph{arXiv preprint arXiv:1312.6114}, 2013.

\bibitem[Kong et~al.(2020)Kong, Ping, Huang, Zhao, and
  Catanzaro]{kong2020diffwave}
Zhifeng Kong, Wei Ping, Jiaji Huang, Kexin Zhao, and Bryan Catanzaro.
\newblock Diffwave: A versatile diffusion model for audio synthesis.
\newblock \emph{arXiv preprint arXiv:2009.09761}, 2020.

\bibitem[Lewis et~al.(2014)Lewis, Anjyo, Rhee, Zhang, Pighin, and
  Deng]{lewis2014practice}
John~P Lewis, Ken Anjyo, Taehyun Rhee, Mengjie Zhang, Frederic~H Pighin, and
  Zhigang Deng.
\newblock Practice and theory of blendshape facial models.
\newblock \emph{Eurographics (State of the Art Reports)}, 1\penalty0
  (8):\penalty0 2, 2014.

\bibitem[Liu et~al.(2022)Liu, Li, Ren, Chen, and Zhao]{liu2022diffsinger}
Jinglin Liu, Chengxi Li, Yi Ren, Feiyang Chen, and Zhou Zhao.
\newblock Diffsinger: Singing voice synthesis via shallow diffusion mechanism.
\newblock In \emph{Proceedings of the AAAI conference on artificial
  intelligence}, pages 11020--11028, 2022.

\bibitem[Loshchilov and Hutter(2017)]{loshchilov2017decoupled}
Ilya Loshchilov and Frank Hutter.
\newblock Decoupled weight decay regularization.
\newblock \emph{arXiv preprint arXiv:1711.05101}, 2017.

\bibitem[Lugmayr et~al.(2022)Lugmayr, Danelljan, Romero, Yu, Timofte, and
  Gool]{Lugmayr2022RePaintIU}
Andreas Lugmayr, Martin Danelljan, Andr{\'e}s Romero, Fisher Yu, Radu Timofte,
  and Luc~Van Gool.
\newblock Repaint: Inpainting using denoising diffusion probabilistic models.
\newblock \emph{2022 IEEE/CVF Conference on Computer Vision and Pattern
  Recognition (CVPR)}, pages 11451--11461, 2022.

\bibitem[Meta(2018)]{meta2018oculus}
Meta.
\newblock Tech note: Enhancing oculus lipsync with deep learning.
\newblock 2018.

\bibitem[Pedregosa et~al.(2011)Pedregosa, Varoquaux, Gramfort, Michel, Thirion,
  Grisel, Blondel, Prettenhofer, Weiss, Dubourg, et~al.]{pedregosa2011scikit}
Fabian Pedregosa, Ga{\"e}l Varoquaux, Alexandre Gramfort, Vincent Michel,
  Bertrand Thirion, Olivier Grisel, Mathieu Blondel, Peter Prettenhofer, Ron
  Weiss, Vincent Dubourg, et~al.
\newblock Scikit-learn: Machine learning in python.
\newblock \emph{the Journal of machine Learning research}, 12:\penalty0
  2825--2830, 2011.

\bibitem[Peng et~al.(2023)Peng, Wu, Song, Xu, Zhu, Liu, He, and
  Fan]{peng2023emotalk}
Ziqiao Peng, Haoyu Wu, Zhenbo Song, Hao Xu, Xiangyu Zhu, Hongyan Liu, Jun He,
  and Zhaoxin Fan.
\newblock Emotalk: Speech-driven emotional disentanglement for 3d face
  animation.
\newblock \emph{arXiv preprint arXiv:2303.11089}, 2023.

\bibitem[Pham et~al.(2017)Pham, Cheung, and Pavlovic]{pham2017speech}
Hai~X Pham, Samuel Cheung, and Vladimir Pavlovic.
\newblock Speech-driven 3d facial animation with implicit emotional awareness:
  A deep learning approach.
\newblock In \emph{Proceedings of the IEEE conference on computer vision and
  pattern recognition workshops}, pages 80--88, 2017.

\bibitem[Pham et~al.(2018)Pham, Wang, and Pavlovic]{pham2018end}
Hai~Xuan Pham, Yuting Wang, and Vladimir Pavlovic.
\newblock End-to-end learning for 3d facial animation from speech.
\newblock In \emph{Proceedings of the 20th ACM International Conference on
  Multimodal Interaction}, pages 361--365, 2018.

\bibitem[Podell et~al.(2023)Podell, English, Lacey, Blattmann, Dockhorn,
  M{\"u}ller, Penna, and Rombach]{podell2023sdxl}
Dustin Podell, Zion English, Kyle Lacey, Andreas Blattmann, Tim Dockhorn, Jonas
  M{\"u}ller, Joe Penna, and Robin Rombach.
\newblock Sdxl: improving latent diffusion models for high-resolution image
  synthesis.
\newblock \emph{arXiv preprint arXiv:2307.01952}, 2023.

\bibitem[Ramesh et~al.(2022)Ramesh, Dhariwal, Nichol, Chu, and
  Chen]{ramesh2022hierarchical}
Aditya Ramesh, Prafulla Dhariwal, Alex Nichol, Casey Chu, and Mark Chen.
\newblock Hierarchical text-conditional image generation with clip latents.
\newblock \emph{arXiv preprint arXiv:2204.06125}, 1\penalty0 (2):\penalty0 3,
  2022.

\bibitem[Richard et~al.(2021{\natexlab{a}})Richard, Lea, Ma, Gall, De~la Torre,
  and Sheikh]{richard2021audio}
Alexander Richard, Colin Lea, Shugao Ma, Jurgen Gall, Fernando De~la Torre, and
  Yaser Sheikh.
\newblock Audio-and gaze-driven facial animation of codec avatars.
\newblock In \emph{Proceedings of the IEEE/CVF winter conference on
  applications of computer vision}, pages 41--50, 2021{\natexlab{a}}.

\bibitem[Richard et~al.(2021{\natexlab{b}})Richard, Zollh{\"o}fer, Wen, De~la
  Torre, and Sheikh]{richard2021meshtalk}
Alexander Richard, Michael Zollh{\"o}fer, Yandong Wen, Fernando De~la Torre,
  and Yaser Sheikh.
\newblock Meshtalk: 3d face animation from speech using cross-modality
  disentanglement.
\newblock In \emph{Proceedings of the IEEE/CVF International Conference on
  Computer Vision}, pages 1173--1182, 2021{\natexlab{b}}.

\bibitem[Rombach et~al.(2022)Rombach, Blattmann, Lorenz, Esser, and
  Ommer]{rombach2022high}
Robin Rombach, Andreas Blattmann, Dominik Lorenz, Patrick Esser, and Bj{\"o}rn
  Ommer.
\newblock High-resolution image synthesis with latent diffusion models.
\newblock In \emph{Proceedings of the IEEE/CVF Conference on Computer Vision
  and Pattern Recognition}, pages 10684--10695, 2022.

\bibitem[Sagonas et~al.(2013)Sagonas, Tzimiropoulos, Zafeiriou, and
  Pantic]{sagonas2013300}
Christos Sagonas, Georgios Tzimiropoulos, Stefanos Zafeiriou, and Maja Pantic.
\newblock 300 faces in-the-wild challenge: The first facial landmark
  localization challenge.
\newblock In \emph{Proceedings of the IEEE international conference on computer
  vision workshops}, pages 397--403, 2013.

\bibitem[Saharia et~al.(2021)Saharia, Chan, Chang, Lee, Ho, Salimans, Fleet,
  and Norouzi]{Saharia2021PaletteID}
Chitwan Saharia, William Chan, Huiwen Chang, Chris~A. Lee, Jonathan Ho, Tim
  Salimans, David~J. Fleet, and Mohammad Norouzi.
\newblock Palette: Image-to-image diffusion models.
\newblock \emph{ACM SIGGRAPH 2022 Conference Proceedings}, 2021.

\bibitem[Saharia et~al.(2022{\natexlab{a}})Saharia, Chan, Saxena, Li, Whang,
  Denton, Ghasemipour, Gontijo~Lopes, Karagol~Ayan, Salimans,
  et~al.]{saharia2022photorealistic}
Chitwan Saharia, William Chan, Saurabh Saxena, Lala Li, Jay Whang, Emily~L
  Denton, Kamyar Ghasemipour, Raphael Gontijo~Lopes, Burcu Karagol~Ayan, Tim
  Salimans, et~al.
\newblock Photorealistic text-to-image diffusion models with deep language
  understanding.
\newblock \emph{Advances in Neural Information Processing Systems},
  35:\penalty0 36479--36494, 2022{\natexlab{a}}.

\bibitem[Saharia et~al.(2022{\natexlab{b}})Saharia, Ho, Chan, Salimans, Fleet,
  and Norouzi]{saharia2022image}
Chitwan Saharia, Jonathan Ho, William Chan, Tim Salimans, David~J Fleet, and
  Mohammad Norouzi.
\newblock Image super-resolution via iterative refinement.
\newblock \emph{IEEE Transactions on Pattern Analysis and Machine
  Intelligence}, 45\penalty0 (4):\penalty0 4713--4726, 2022{\natexlab{b}}.

\bibitem[Shafir et~al.(2023)Shafir, Tevet, Kapon, and Bermano]{shafir2023human}
Yonatan Shafir, Guy Tevet, Roy Kapon, and Amit~H Bermano.
\newblock Human motion diffusion as a generative prior.
\newblock \emph{arXiv preprint arXiv:2303.01418}, 2023.

\bibitem[Sohl-Dickstein et~al.(2015)Sohl-Dickstein, Weiss, Maheswaranathan, and
  Ganguli]{sohl2015deep}
Jascha Sohl-Dickstein, Eric Weiss, Niru Maheswaranathan, and Surya Ganguli.
\newblock Deep unsupervised learning using nonequilibrium thermodynamics.
\newblock In \emph{International Conference on Machine Learning}, pages
  2256--2265. PMLR, 2015.

\bibitem[Song et~al.(2020)Song, Meng, and Ermon]{Song2020DenoisingDI}
Jiaming Song, Chenlin Meng, and Stefano Ermon.
\newblock Denoising diffusion implicit models.
\newblock \emph{ArXiv}, abs/2010.02502, 2020.

\bibitem[Stan et~al.(2023)Stan, Haque, and Yumak]{stan2023facediffuser}
Stefan Stan, Kazi~Injamamul Haque, and Zerrin Yumak.
\newblock Facediffuser: Speech-driven 3d facial animation synthesis using
  diffusion.
\newblock \emph{arXiv preprint arXiv:2309.11306}, 2023.

\bibitem[Sumner and Popovi{\'c}(2004)]{sumner2004deformation}
Robert~W Sumner and Jovan Popovi{\'c}.
\newblock Deformation transfer for triangle meshes.
\newblock \emph{ACM Transactions on graphics (TOG)}, 23\penalty0 (3):\penalty0
  399--405, 2004.

\bibitem[Tarvainen and Valpola(2017)]{tarvainen2017mean}
Antti Tarvainen and Harri Valpola.
\newblock Mean teachers are better role models: Weight-averaged consistency
  targets improve semi-supervised deep learning results.
\newblock \emph{Advances in neural information processing systems}, 30, 2017.

\bibitem[Taylor et~al.(2017)Taylor, Kim, Yue, Mahler, Krahe, Rodriguez,
  Hodgins, and Matthews]{taylor2017deep}
Sarah Taylor, Taehwan Kim, Yisong Yue, Moshe Mahler, James Krahe,
  Anastasio~Garcia Rodriguez, Jessica Hodgins, and Iain Matthews.
\newblock A deep learning approach for generalized speech animation.
\newblock \emph{ACM Transactions On Graphics (TOG)}, 36\penalty0 (4):\penalty0
  1--11, 2017.

\bibitem[Taylor et~al.(2012)Taylor, Mahler, Theobald, and
  Matthews]{taylor2012dynamic}
Sarah~L Taylor, Moshe Mahler, Barry-John Theobald, and Iain Matthews.
\newblock Dynamic units of visual speech.
\newblock In \emph{Proceedings of the 11th ACM SIGGRAPH/Eurographics conference
  on Computer Animation}, pages 275--284, 2012.

\bibitem[Tevet et~al.(2023)Tevet, Raab, Gordon, Shafir, Cohen-or, and
  Bermano]{tevet2023human}
Guy Tevet, Sigal Raab, Brian Gordon, Yoni Shafir, Daniel Cohen-or, and
  Amit~Haim Bermano.
\newblock Human motion diffusion model.
\newblock In \emph{The Eleventh International Conference on Learning
  Representations}, 2023.

\bibitem[Tseng et~al.(2023)Tseng, Castellon, and Liu]{tseng2023edge}
Jonathan Tseng, Rodrigo Castellon, and Karen Liu.
\newblock Edge: Editable dance generation from music.
\newblock In \emph{Proceedings of the IEEE/CVF Conference on Computer Vision
  and Pattern Recognition}, pages 448--458, 2023.

\bibitem[Vaswani et~al.(2017)Vaswani, Shazeer, Parmar, Uszkoreit, Jones, Gomez,
  Kaiser, and Polosukhin]{vaswani2017attention}
Ashish Vaswani, Noam Shazeer, Niki Parmar, Jakob Uszkoreit, Llion Jones,
  Aidan~N Gomez, {\L}ukasz Kaiser, and Illia Polosukhin.
\newblock Attention is all you need.
\newblock \emph{Advances in neural information processing systems}, 30, 2017.

\bibitem[Xing et~al.(2023)Xing, Xia, Zhang, Cun, Wang, and
  Wong]{xing2023codetalker}
Jinbo Xing, Menghan Xia, Yuechen Zhang, Xiaodong Cun, Jue Wang, and Tien-Tsin
  Wong.
\newblock Codetalker: Speech-driven 3d facial animation with discrete motion
  prior.
\newblock \emph{arXiv preprint arXiv:2301.02379}, 2023.

\bibitem[Xu et~al.(2013)Xu, Feng, Marsella, and Shapiro]{xu2013practical}
Yuyu Xu, Andrew~W Feng, Stacy Marsella, and Ari Shapiro.
\newblock A practical and configurable lip sync method for games.
\newblock In \emph{Proceedings of Motion on Games}, pages 131--140. 2013.

\bibitem[Yoon et~al.(2020)Yoon, Cha, Lee, Jang, Lee, Kim, and
  Lee]{yoon2020speech}
Youngwoo Yoon, Bok Cha, Joo-Haeng Lee, Minsu Jang, Jaeyeon Lee, Jaehong Kim,
  and Geehyuk Lee.
\newblock Speech gesture generation from the trimodal context of text, audio,
  and speaker identity.
\newblock \emph{ACM Transactions on Graphics (TOG)}, 39\penalty0 (6):\penalty0
  1--16, 2020.

\bibitem[Yuan et~al.(2022)Yuan, Song, Iqbal, Vahdat, and
  Kautz]{yuan2022physdiff}
Ye Yuan, Jiaming Song, Umar Iqbal, Arash Vahdat, and Jan Kautz.
\newblock Physdiff: Physics-guided human motion diffusion model.
\newblock \emph{arXiv preprint arXiv:2212.02500}, 2022.

\bibitem[Zhang et~al.(2022)Zhang, Cai, Pan, Hong, Guo, Yang, and
  Liu]{zhang2022motiondiffuse}
Mingyuan Zhang, Zhongang Cai, Liang Pan, Fangzhou Hong, Xinying Guo, Lei Yang,
  and Ziwei Liu.
\newblock Motiondiffuse: Text-driven human motion generation with diffusion
  model.
\newblock \emph{arXiv preprint arXiv:2208.15001}, 2022.

\bibitem[Zhou et~al.(2018)Zhou, Xu, Landreth, Kalogerakis, Maji, and
  Singh]{zhou2018visemenet}
Yang Zhou, Zhan Xu, Chris Landreth, Evangelos Kalogerakis, Subhransu Maji, and
  Karan Singh.
\newblock Visemenet: Audio-driven animator-centric speech animation.
\newblock \emph{ACM Transactions on Graphics (TOG)}, 37\penalty0 (4):\penalty0
  1--10, 2018.

\bibitem[Zhu et~al.(2023)Zhu, Liu, Liu, Qian, Liu, and Yu]{Zhu_2023_CVPR}
Lingting Zhu, Xian Liu, Xuanyu Liu, Rui Qian, Ziwei Liu, and Lequan Yu.
\newblock Taming diffusion models for audio-driven co-speech gesture
  generation.
\newblock In \emph{Proceedings of the IEEE/CVF Conference on Computer Vision
  and Pattern Recognition (CVPR)}, pages 10544--10553, 2023.

\end{thebibliography}
}

\clearpage
\setcounter{page}{1}
\maketitlesupplementary

\section{Equivalent Form of \cref{eq:qp}}\label{app:qp}

\cref{eq:qp} is equivalent to the following formulation:
\begin{align}
\begin{split}
\min_{\bm{u}^{1:N}} \quad & \left\lVert
\begin{bmatrix}
\bm{p}^{1} \\ \bm{p}^{2} \\ \vdots \\ \bm{p}^{N}
\end{bmatrix}
 - \left(
\begin{bmatrix}
\bm{b}_0 \\ \bm{b}_0 \\ \vdots \\ \bm{b}_0
\end{bmatrix}
+
\begin{bmatrix}
\bm{B} & \bm{0} & \cdots & \bm{0} \\
\bm{0} & \bm{B} & \cdots & \bm{0} \\
\vdots & \vdots & \ddots & \bm{0} \\
\bm{0} & \bm{0} & \cdots & \bm{B} \\
\end{bmatrix}
\begin{bmatrix}
\bm{u}^{1} \\ \bm{u}^{2} \\ \vdots \\ \bm{u}^{N}
\end{bmatrix}
\right) \right\rVert_{2}^{2} \\
\textrm{s.t.}
\quad &\bm{0} \preceq
\begin{bmatrix}
\bm{u}^{1} \\ \bm{u}^{2} \\ \vdots \\ \bm{u}^{N}
\end{bmatrix}
\preceq \bm{1},\\
\quad &
\begin{bmatrix}
\bm{I} & -\bm{I} & \bm{0} & \cdots & \bm{0} \\
\bm{0} & \bm{I} & -\bm{I} & \cdots & \bm{0} \\
\vdots & \vdots & \ddots & \ddots & \bm{0} \\
\bm{0} & \bm{0} & \cdots & \bm{I} & -\bm{I} \\
\end{bmatrix}
\begin{bmatrix}
\bm{u}^{1} \\ \bm{u}^{2} \\ \vdots \\ \bm{u}^{N}
\end{bmatrix}
\preceq \delta \cdot \bm{1},\\
\quad &
\begin{bmatrix}
-\bm{I} & \bm{I} & \bm{0} & \cdots & \bm{0} \\
\bm{0} & -\bm{I} & \bm{I} & \cdots & \bm{0} \\
\vdots & \vdots & \ddots & \ddots & \bm{0} \\
\bm{0} & \bm{0} & \cdots & -\bm{I} & \bm{I} \\
\end{bmatrix}
\begin{bmatrix}
\bm{u}^{1} \\ \bm{u}^{2} \\ \vdots \\ \bm{u}^{N}
\end{bmatrix}
\preceq \delta \cdot \bm{1},\\
\end{split}
\end{align}
where $\preceq$ denotes the element-wise comparison operator and $\bm{B} = [\bm{b}_1 - \bm{b}_0 | \bm{b}_2 - \bm{b}_0 | \cdots | \bm{b}_K - \bm{b}_0 ]$ is a matrix whose column vectors are residual blendshape position vectors.
We can simplify the problem as follows:
\begin{align}\label{appendix:prob:qp}
\begin{split}
\min_{\bm{u}} \quad& \frac{1}{2} {\bm{u}}^{\intercal} \bm{P} \bm{u} + \bm{q}^{\intercal} \bm{u}
\quad \textrm{s.t.}
\quad \bm{0} \preceq \bm{u} \preceq \bm{1},
\quad \bm{G} \bm{u} \preceq \delta \cdot \bm{1},
\end{split}
\end{align}
where
\begin{align*}
\bm{u} &=
\begin{bmatrix}
\bm{u}^{1} \\ \bm{u}^{2} \\ \vdots \\ \bm{u}^{N}
\end{bmatrix}, 
\quad \bm{P} =
\begin{bmatrix}
\bm{B}^{\intercal} \bm{B} & \bm{0} & \cdots & \bm{0} \\
\bm{0} & \bm{B}^{\intercal} \bm{B} & \cdots & \bm{0} \\
\vdots & \vdots & \ddots & \bm{0} \\
\bm{0} & \bm{0} & \cdots & \bm{B}^{\intercal} \bm{B} \\
\end{bmatrix}, \\
\quad \bm{q} &=
\begin{bmatrix}
\bm{B}^{\intercal} ( \bm{b}_0 - \bm{p}^{1} ) \\
\bm{B}^{\intercal} ( \bm{b}_0 - \bm{p}^{2} ) \\
\vdots \\
\bm{B}^{\intercal} ( \bm{b}_0 - \bm{p}^{N} ) \\
\end{bmatrix},
\quad \bm{D} =
\begin{bmatrix}
\bm{I} \\
-\bm{I}
\end{bmatrix},\\
\bm{G} &=
\begin{bmatrix}
\bm{D} & -\bm{D} & \bm{0} & \cdots & \bm{0} \\
\bm{0} & \bm{D} & -\bm{D} & \cdots & \bm{0} \\
\vdots & \vdots & \ddots & \ddots & \vdots \\
\bm{0} & \bm{0} & \cdots & \bm{D} & -\bm{D} \\
\end{bmatrix}.
\end{align*}
As we can see, the objective function is convex quadratic, and every constraint function is affine.
Therefore, it is a quadratic program.

In most case, $\bm{b}_0, \bm{b}_1, \cdots, \bm{b}_K$ are designed to satisfy $\lVert \bm{B} \bm{u}^{n} \rVert_{2} > 0$ for any $\bm{u}^{n} \neq \bm{0}$ to increase the degree of freedom of the blendshape facial model.
Therefore, $\bm{B}^{\intercal} \bm{B}$ is a positive definite matrix.
Consequently, $\bm{P}$ is also a positive definite, since ${\bm{u}}^{\intercal} \bm{P} \bm{u} = \sum_{n=1}^{N} {\bm{u}^{n}}^{\intercal} (\bm{B}^{\intercal} \bm{B}) \bm{u}^{n} > 0$ holds for arbitrary $\bm{u} \neq \bm{0}$.
As a result, the objective in \cref{appendix:prob:qp} is strictly convex, which means that the optimization problem has a unique solution.

\section{Baseline Methods}\label{app:baseline}

We compare SAiD with end2end\_AU\_speech~\citep{pham2018end}, which regresses the blendshape coefficients given the audio spectrogram.
We use CNN+GRU configuration, which results in the lowest errors among the configurations that handle the smooth temporal transition.
To prevent overfitting, we modify the number of convolution layers of the audio encoder from 8 to 5 by changing the convolution layers into the max pooling layers.
We modify the convolution layers into max pooling layer since 
We train and test end2end\_AU\_speech on the dataset described in \cref{sec:method:dataset}.

Considering the limited available models for speech-driven blendshape facial animation, we further compare SAiD with the models that directly produce the mesh sequence of the facial motion.
We select four state-of-the-art methods, VOCA~\citep{cudeiro2019capture}, MeshTalk~\citep{richard2021meshtalk}, FaceFormer~\citep{fan2022faceformer}, and CodeTalker~\citep{xing2023codetalker} as baselines.
We use the pre-trained weights of each model to obtain the predictions of baselines.
Exceptionally, we train and test MeshTalk on VOCASET with the modification suggested by \citet{fan2022faceformer}.
Next, we apply the linear interpolation on the outputs of FaceFormer and CodeTalker to change the frame rate from 30fps to 60fps.
After that, we optimize the blendshape coefficient sequence from the generated mesh sequence by solving the QP problem described in \cref{sec:method:dataset:qp}.

\section{Evaluation Metrics}\label{app:eval_metric}

\paragraph{Audio-visual offset/confidence~\citep{chung2017out}:} {
        AV offset and confidence check the synchronization offset and confidence between the audio and the video using SyncNet~\citep{chung2017out}.
        We use the rendered videos of the frontal view of the reconstructed mesh sequences.
        Next, we report the mean offset and mean confidence among the videos.
    }
    \vspace{-4mm}
    \paragraph{Multimodality~\citep{guo2020action2motion}:} {
        Multimodality measures how much the generated animations diversify given audio.
        Given a set of $C$ audio clips, we randomly sample two subsets of blendshape coefficient sequences with the same size $S_l$ for each audio clip.
        Next, we extract two subsets of latent features $\{\bm{v}_{c,1}, \cdots, \bm{v}_{c,S_l}\}$ and $\{\bm{v}_{c,1}', \cdots, \bm{v}_{c,S_l}'\}$.
        Then, the multimodality is defined as follows:
        \begin{align}
            \mathrm{Multimodality} = \frac{1}{C \times S_l} \sum_{c=1}^{C} \sum_{i=1}^{S_l} \lVert \bm{v}_{c,i} - \bm{v}_{c,i}' \rVert_{2}.
        \end{align}
        We set $S_l = 36$ for the evaluation.
    }
    \vspace{-4mm}
    \paragraph{Fr\'echet distance (FD)~\citep{dowson1982frechet, heusel2017gans}:} {
        We use FD between the the latent features of the real and the generated blendshape coefficient sequences:
        \begin{align}
        \begin{split}
            \mathrm{FD} \left(P_r, P_g\right) &= \lVert \bm{\mu}_r - \bm{\mu}_g \rVert_2^2\\ &+ \mathrm{Tr} \left( \bm{\Sigma}_r + \bm{\Sigma}_g - 2 \left(\bm{\Sigma}_r \bm{\Sigma}_g \right)^{1/2} \right),
        \end{split}
        \end{align}
        where $P_r = \mathcal{N} (\bm{\mu}_r, \bm{\Sigma}_r)$ and $P_g = \mathcal{N} (\bm{\mu}_g, \bm{\Sigma}_g)$ are the latent distributions of real/generated blendshape coefficient sequence, respectively.
    }
    \vspace{-4mm}
    \paragraph{Wasserstein inception distance (WInD)~\citep{dimitrakopoulos2020wind}:} {
        WInD is an extended version of FD, assuming that real and generated latent distribution follow the Gaussian Mixture Model (GMM).
        We can compute it as a result of the following linear programming (LP) problem:
        \begin{align}
        \begin{split}
            &\mathrm{WInD} \left(P_r, P_g\right) = \min_{w^{ij} \ge 0} \sum_{i=1}^{K} \sum_{j=1}^{K} w^{ij} d(i,j)\\
            &\textrm{s.t.}~
            \sum_{i=1}^{K} w^{ij} \le \pi^{j},
            \sum_{j=1}^{K} w^{ij} \le \pi^{i},
            \sum_{i=1}^{K}\sum_{j=1}^{K} w^{ij} = 1,
        \end{split}
        \end{align}
        where $P_r = \sum_{i=1}^{K} \pi^{i} \mathcal{N} (\bm{\mu}_r^{i}, \bm{\Sigma}_r^{i})$ and $P_g = \sum_{j=1}^{K} \pi^{j} \mathcal{N} (\bm{\mu}_g^{j}, \bm{\Sigma}_g^{j})$ are the latent distributions of real/generated blendshape coefficient sequences and $d(i, j)$ is a Wasserstein distance between two gaussian distributions $\mathcal{N} (\bm{\mu}_r^{i}, \bm{\Sigma}_r^{i}) $ and $\mathcal{N} (\bm{\mu}_g^{j}, \bm{\Sigma}_g^{j})$.
        We use Scikit-learn~\citep{pedregosa2011scikit} to get GMM with $K=5$ and CVXOPT~\citep{andersen2013cvxopt} to solve the LP problem.
        Since WInD can be varied depending on the constructed GMMs, we repeat the computation $10$ times and report the mean and standard deviation.
    }

Since there is no previous feature extractor for blendshape coefficient sequences, we train the variational autoencoder (VAE)~\citep{kingma2013auto} using the same training dataset in \cref{sec:experiment:training}.
We use a latent mean of VAE as a latent feature when computing the evaluation metrics (FD, WInD, multimodality).
The architecture and the training details are in \cref{app:feature_extractor}.

\section{Feature Extractor for Blendshape Coefficient Sequence}\label{app:feature_extractor}

\subsection{Model}

\begin{figure}[t]
    \centering
    \includegraphics[width=\linewidth]{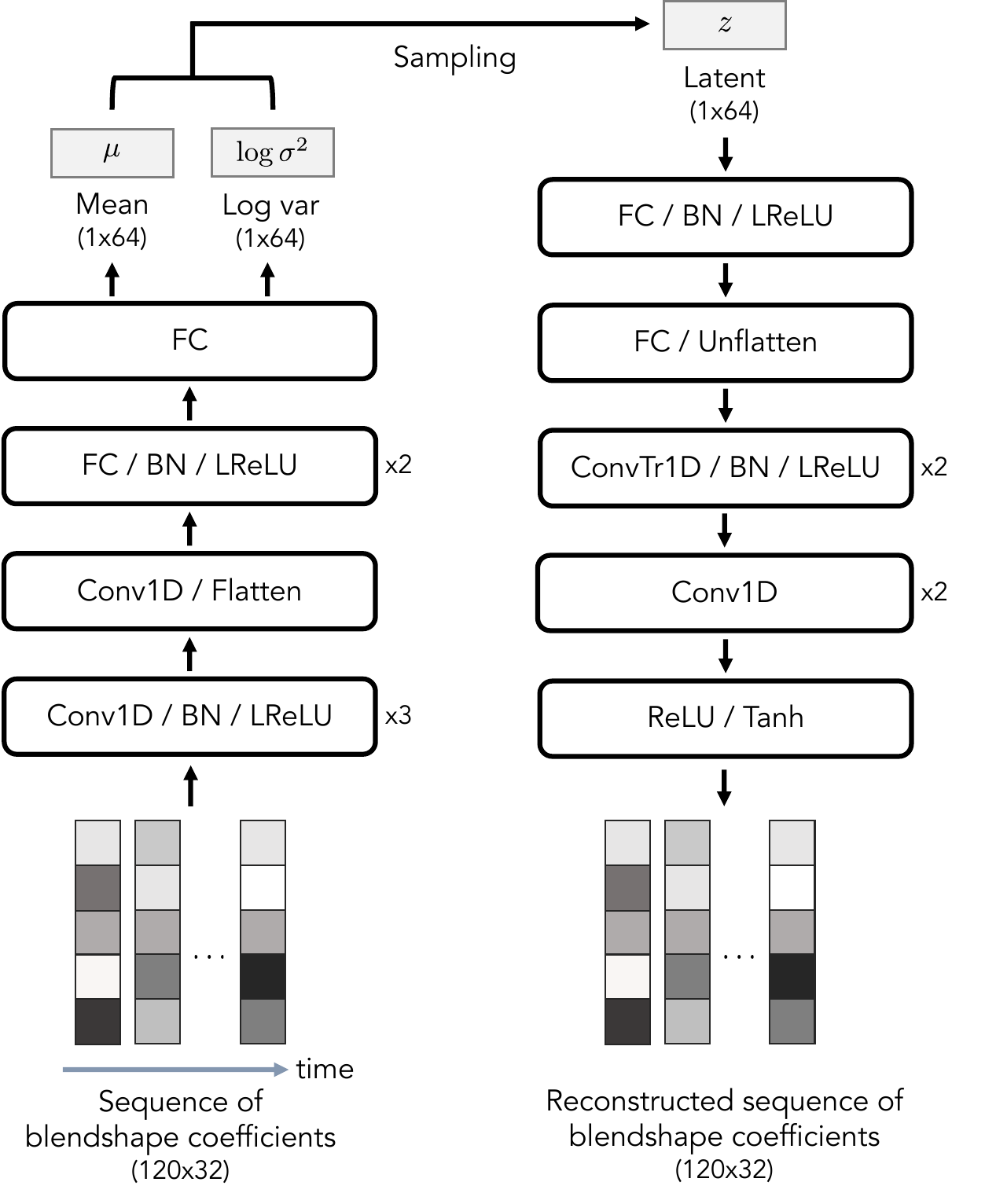}
    \caption{
        \textbf{The model architecture of VAE.}
    }
    \label{fig:vae}
\end{figure}

We utilize the encoder part of the VAE as a feature extractor for the blendshape coefficient sequences.
As illustrated in \cref{fig:vae}, VAE structure is adapted from~\citet{yoon2020speech}, with modifications in the model size. Additionally, we add ReLU and Tanh layers at the end of the decoder.

\subsection{Training Details}

We train the VAE by minimizing the reconstruction error with the regularization term.
We measure the reconstruction error using the weighted squared error between the input and its corresponding output.
The regularization term represents the KL divergence between the latent and standard normal distributions.
\begin{align}
\begin{split}
    &\mathcal{L}_{\mathrm{reconst}} (\bm{\theta}, \bm{\phi}) = \mathbb{E}_{ q}  \Bigl[ \lVert \mathrm{diag} (\bm{\sigma}_{\bm{u}})^{-1} (\bm{u}^{1:N} - \hat{\bm{u}}^{1:N} (\bm{\theta}, \bm{\phi}) ) \rVert_F^2 \Bigr],\\
    &\mathcal{L}_{\mathrm{reg}} (\bm{\phi}) =\\
    &~~\mathbb{E}_{ q} \Bigl[  \mathcal{D}_{\mathrm{KL}} \left( \mathcal{N} (\bm{\mu}_{\bm{\phi}} (\bm{u}^{1:N}) ), \mathrm{diag} ( \bm{\sigma}_{\bm{\phi}} (\bm{u}^{1:N}) )^2 ) \middle\| \mathcal{N} (\bm{0}, \bm{I}) \right)  \Bigr] ,
\end{split}
\end{align}
where $\bm{\phi}$ and $\bm{\theta}$ are the model parameters of the encoder and decoder part, respectively.
$\bm{\sigma}_{\bm{u}}$ denotes the standard deviation of the blendshape coefficient.
$\hat{\bm{u}}^{1:N} (\bm{\theta}, \bm{\phi})$ is the reconstructed output of the input $\bm{u}^{1:N}$.
$\bm{\mu}_{\bm{\phi}} (\bm{u}^{1:N})$ and $\bm{\sigma}_{\bm{\phi}} (\bm{u}^{1:N})$ are the mean and standard deviation of the latent.

In addition, we use velocity loss to reduce the jitter of the output, which is defined as the gap between the temporal differences of and $\bm{u}^{1:N}$ and $\hat{\bm{u}}^{1:N} = \hat{\bm{u}}^{1:N} (\bm{\theta}, \bm{\phi})$.
\begin{align}
\begin{split}
    &\mathcal{L}_{\mathrm{vel}} (\bm{\theta}, \bm{\phi}) =\\
    &~~\mathbb{E}_{ q}  \Bigl[ \sum_{n=1}^{N-1} \lVert \mathrm{diag} (\bm{\sigma}_{\bm{u}})^{-1} ((\bm{u}^{n+1}-\bm{u}^{n}) - (\hat{\bm{u}}^{n+1} - \hat{\bm{u}}^{n}) ) \rVert_F^2 \Bigr].
\end{split}
\end{align}

As a result, our training loss is:
\begin{align}
\begin{split}
    \mathcal{L}_{\mathrm{VAE}} (\bm{\theta}, \bm{\phi}) = \mathcal{L}_{\mathrm{reconst}} (\bm{\theta}, \bm{\phi}) + \mathcal{L}_{\mathrm{vel}} (\bm{\theta}, \bm{\phi}) + \beta \mathcal{L}_{\mathrm{reg}} (\bm{\phi}),
\end{split}
\end{align}
where $\beta$ is a weighting hyperparameter that follows cyclical annealing schedule~\citep{Fu2019CyclicalAS}.

We use the same training/validation/test dataset for training SAiD as explained in \cref{sec:experiment:training}.
For each training step, we randomly choose a minibatch of size $8$.
Each data in the minibatch consists of the randomly sliced blendshape coefficient sequence.

We conduct the training of VAE on a single NVIDIA A100 (40GB) for 100,000 epochs using the AdamW~\citep{loshchilov2017decoupled} optimizer with $\beta_1=0.9$, $\beta_2 = 0.999$, a learning rate of $10^{-4}$ with warmup, and a weight decay of $10^{-2}$.
We use EMA~\citep{tarvainen2017mean} with a decay value of $0.99$ to update the model weights.

\section{Experimental Settings}\label{app:exp_setting}

We have established the following settings to provide a consistent evaluation environment for all methods:
\vspace{-4mm}
\paragraph{SAiD:} {
    We use a DDIM sampler with $1000$ sampling steps and the guidance strength $\gamma=2.0$.
    We generate $72$ random blendshape coefficient sequences for each audio.
}
\vspace{-4mm}
\paragraph{end2end\_AU\_speech, MeshTalk:} {
    Since these methods are regression-based models without conditions, we generate one unique sequence for each audio.
}
\vspace{-4mm}
\paragraph{VOCA, FaceFormer, CodeTalker:} {
    We generate $8$ different sequences for each audio by conditioning on all training speaker styles.
    To compute the multimodality, we instead use the average distance among every $2$-combination with repetition of latent features, resulting in $\binom{8+2-1}{2} = 36$ pairs for each audio.
}

These conditions ensure that each method is evaluated in as uniform a way as possible.

\begin{figure*}[t]
    \centering
    \captionsetup[subfigure]{labelformat=empty, justification=centering}
    \begin{subfigure}[t]{0.13\linewidth}
        \centering
        \includegraphics[width=\linewidth]{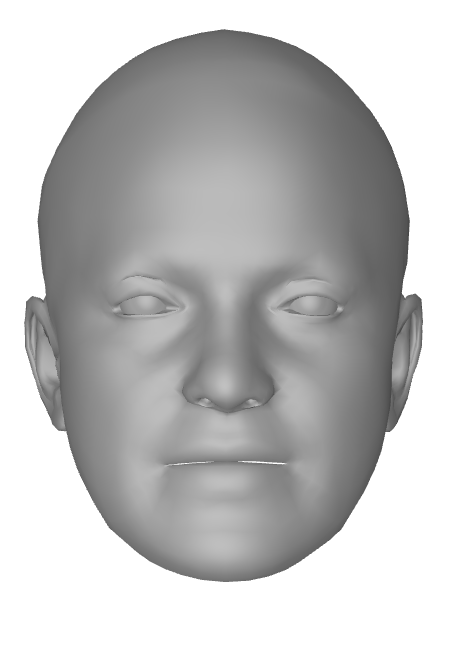}
        \caption{\textbf{Template}}
    \end{subfigure}
    \begin{subfigure}[t]{0.13\linewidth}
        \centering
        \includegraphics[width=\linewidth]{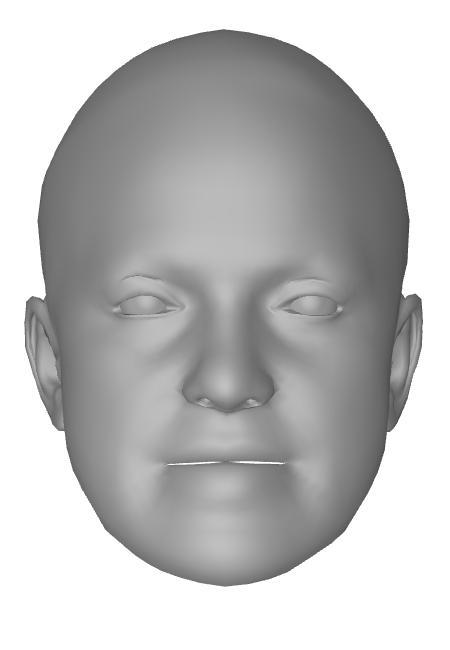}
        \caption{jawForward}
    \end{subfigure}
    \begin{subfigure}[t]{0.13\linewidth}
        \centering
        \includegraphics[width=\linewidth]{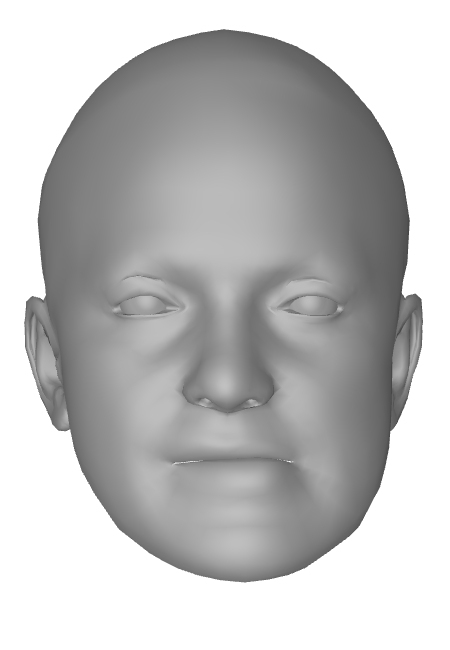}
        \caption{jawLeft}
    \end{subfigure}
    \begin{subfigure}[t]{0.13\linewidth}
        \centering
        \includegraphics[width=\linewidth]{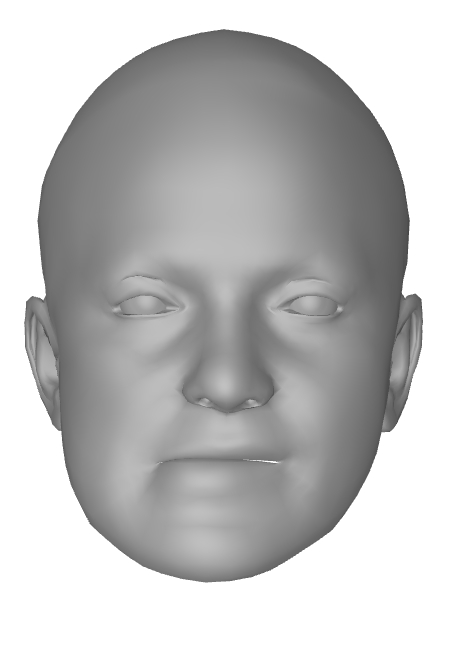}
        \caption{jawRight}
    \end{subfigure}
    \begin{subfigure}[t]{0.13\linewidth}
        \centering
        \includegraphics[width=\linewidth]{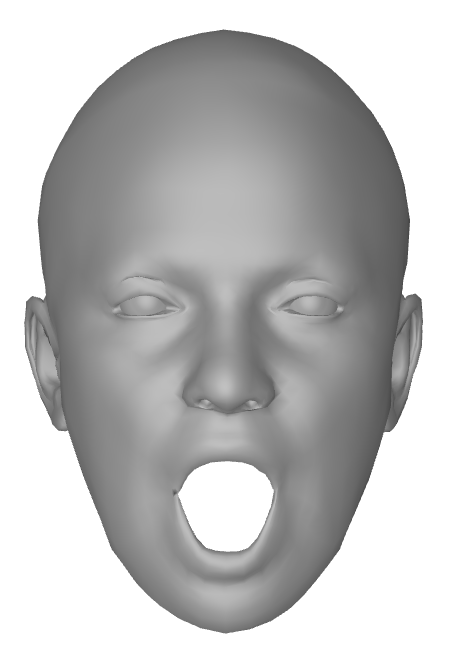}
        \caption{jawOpen}
    \end{subfigure}
    \begin{subfigure}[t]{0.13\linewidth}
        \centering
        \includegraphics[width=\linewidth]{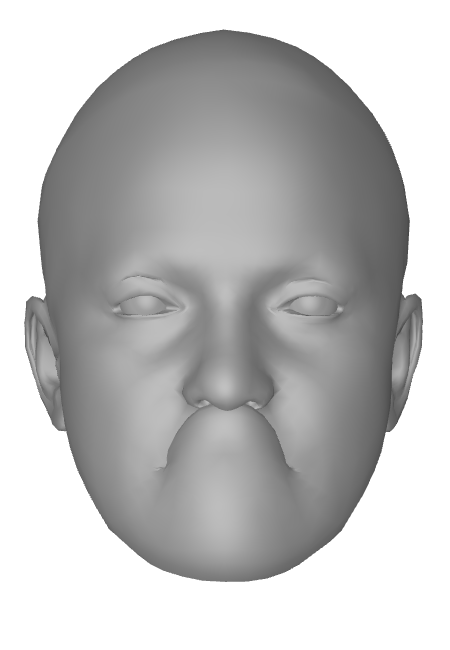}
        \caption{mouthClose}
    \end{subfigure}
    \begin{subfigure}[t]{0.13\linewidth}
        \centering
        \includegraphics[width=\linewidth]{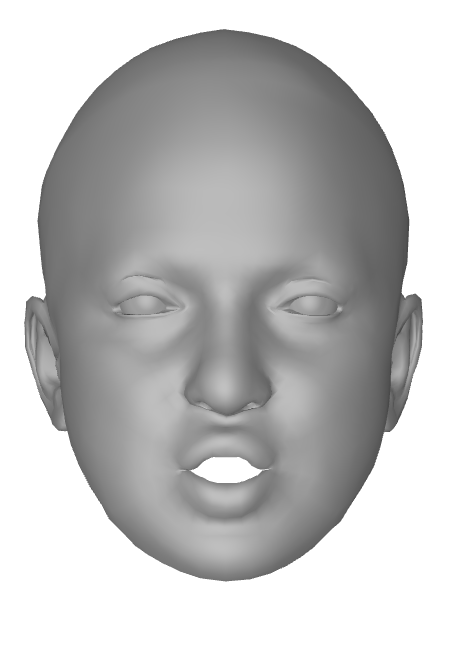}
        \caption{mouthFunnel}
    \end{subfigure}
    \begin{subfigure}[t]{0.13\linewidth}
        \centering
        \includegraphics[width=\linewidth]{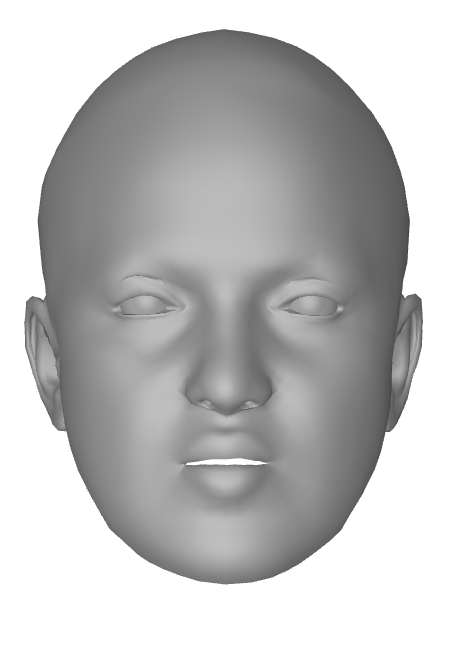}
        \caption{mouthPucker}
    \end{subfigure}
    \begin{subfigure}[t]{0.13\linewidth}
        \centering
        \includegraphics[width=\linewidth]{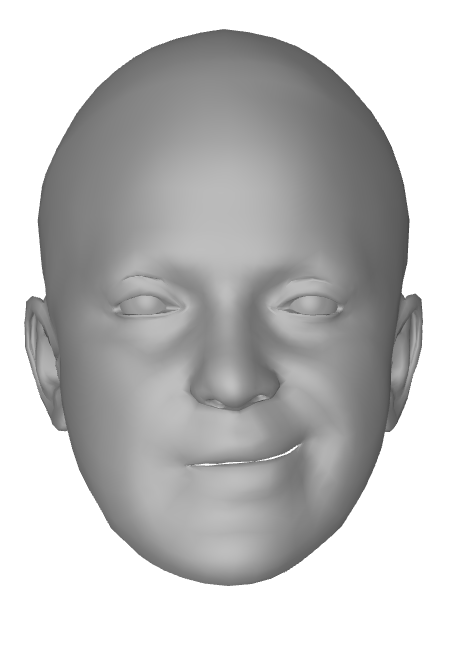}
        \caption{mouthLeft}
    \end{subfigure}
    \begin{subfigure}[t]{0.13\linewidth}
        \centering
        \includegraphics[width=\linewidth]{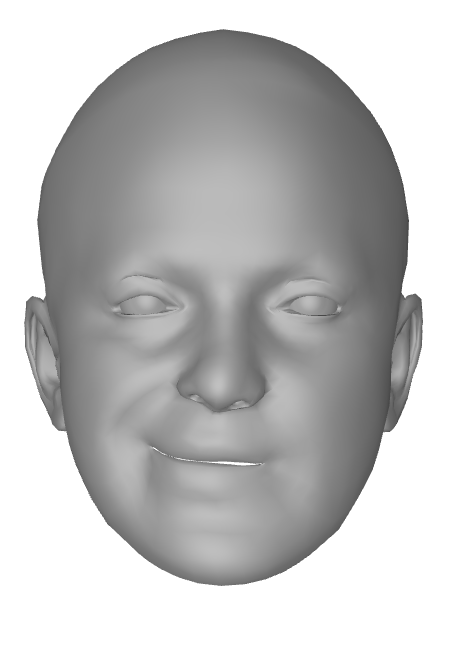}
        \caption{mouthRight}
    \end{subfigure}
    \begin{subfigure}[t]{0.13\linewidth}
        \centering
        \includegraphics[width=\linewidth]{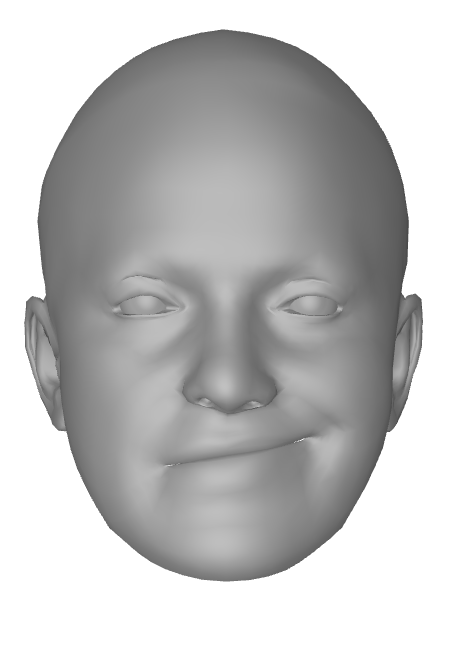}
        \caption{mouth\\SmileLeft}
    \end{subfigure}
    \begin{subfigure}[t]{0.13\linewidth}
        \centering
        \includegraphics[width=\linewidth]{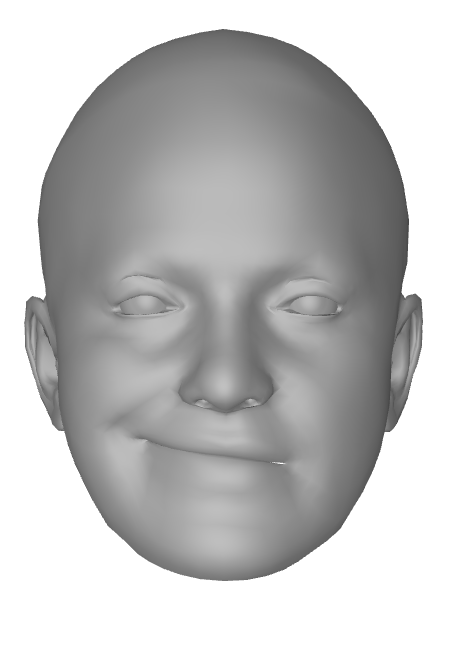}
        \caption{mouth\\SmileRight}
    \end{subfigure}
    \begin{subfigure}[t]{0.13\linewidth}
        \centering
        \includegraphics[width=\linewidth]{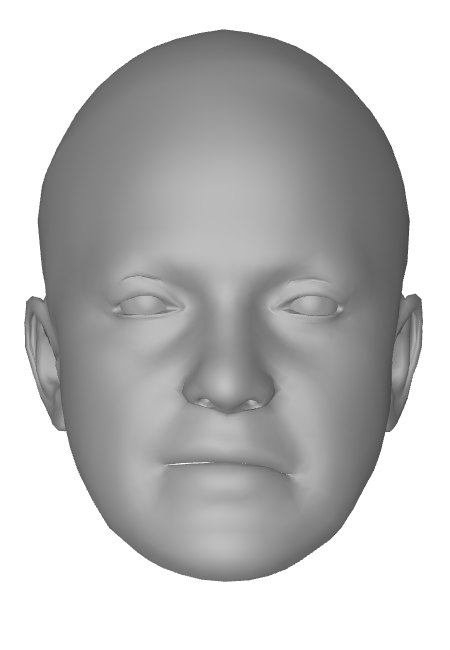}
        \caption{mouth\\FrownLeft}
    \end{subfigure}
    \begin{subfigure}[t]{0.13\linewidth}
        \centering
        \includegraphics[width=\linewidth]{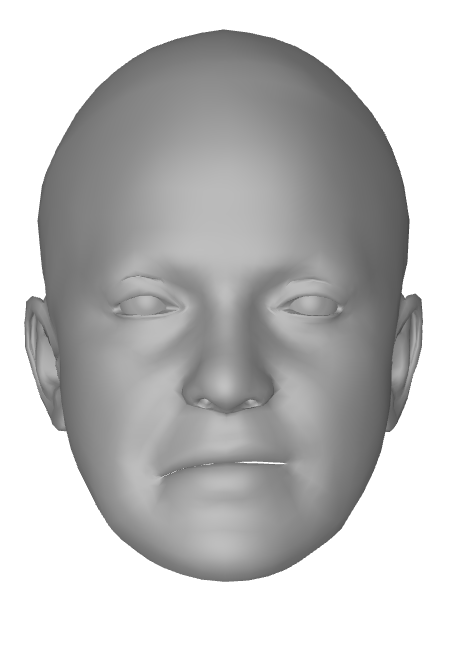}
        \caption{mouth\\FrownRight}
    \end{subfigure}
    \begin{subfigure}[t]{0.13\linewidth}
        \centering
        \includegraphics[width=\linewidth]{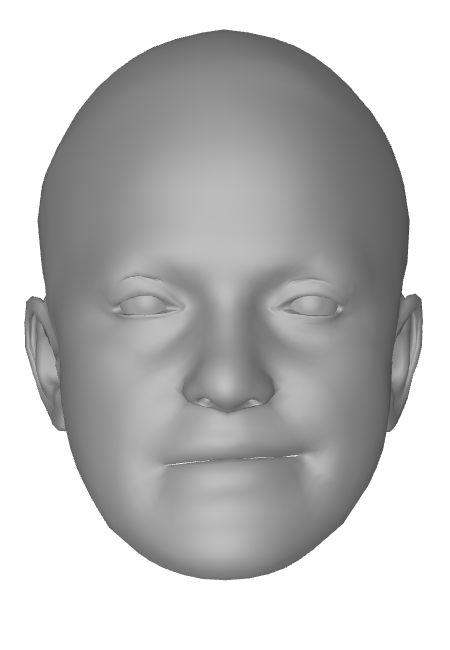}
        \caption{mouth\\DimpleLeft}
    \end{subfigure}
    \begin{subfigure}[t]{0.13\linewidth}
        \centering
        \includegraphics[width=\linewidth]{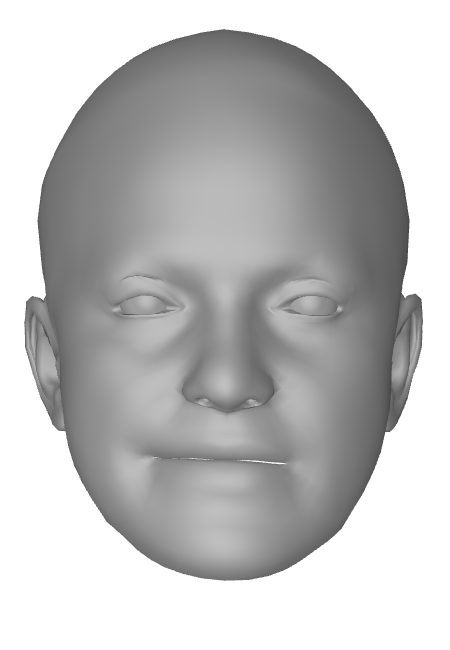}
        \caption{mouth\\DimpleRight}
    \end{subfigure}
    \begin{subfigure}[t]{0.13\linewidth}
        \centering
        \includegraphics[width=\linewidth]{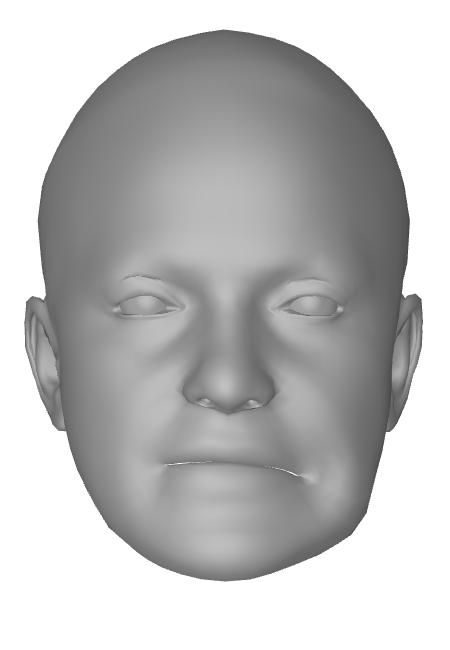}
        \caption{mouth\\StretchLeft}
    \end{subfigure}
    \begin{subfigure}[t]{0.13\linewidth}
        \centering
        \includegraphics[width=\linewidth]{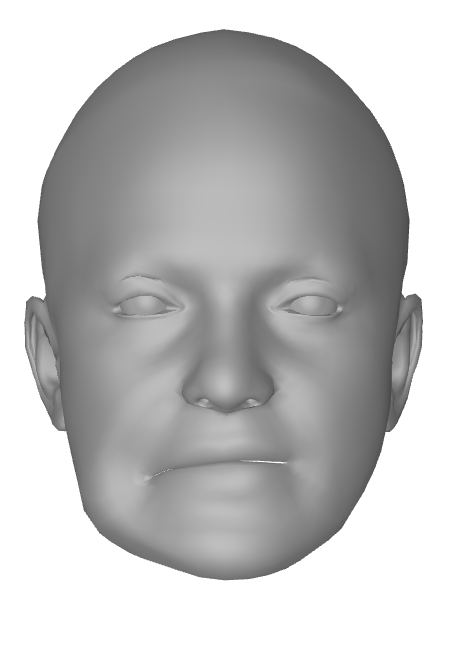}
        \caption{mouth\\StretchRight}
    \end{subfigure}
    \begin{subfigure}[t]{0.13\linewidth}
        \centering
        \includegraphics[width=\linewidth]{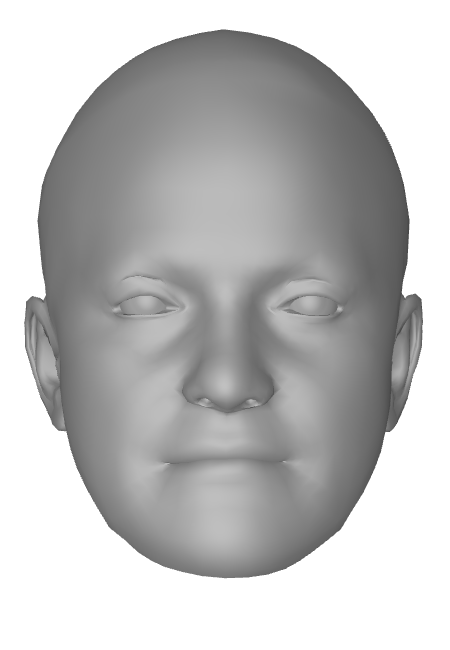}
        \caption{mouth\\RollLower}
    \end{subfigure}
    \begin{subfigure}[t]{0.13\linewidth}
        \centering
        \includegraphics[width=\linewidth]{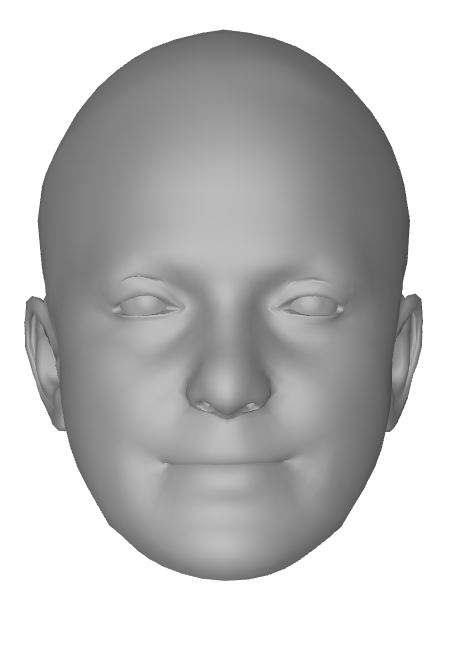}
        \caption{mouth\\RollUpper}
    \end{subfigure}
    \begin{subfigure}[t]{0.13\linewidth}
        \centering
        \includegraphics[width=\linewidth]{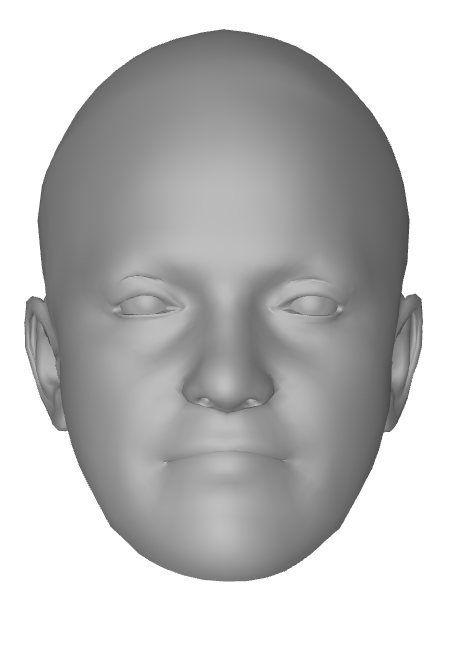}
        \caption{mouth\\ShrugLower}
    \end{subfigure}
\end{figure*}
\begin{figure*}[t]  % 한 figure에 많은 수의 subcaption이 포함될 경우 오류 발생
    \vspace*{-0.7in}
    \addtocounter{figure}{-1}
    \centering
    \captionsetup[subfigure]{labelformat=empty, justification=centering}
    \begin{subfigure}[t]{0.13\linewidth}
        \centering
        \includegraphics[width=\linewidth]{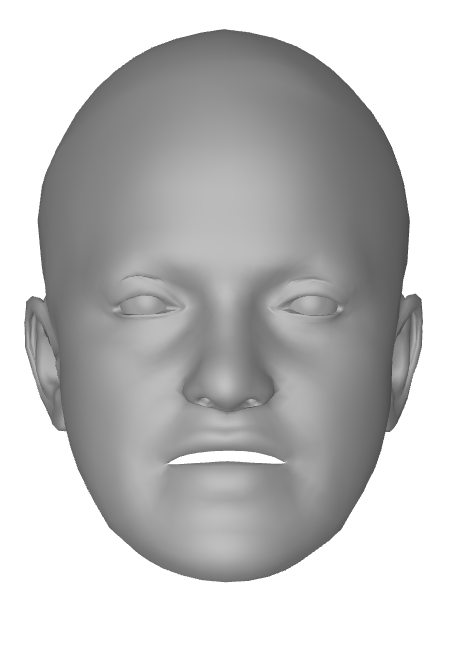}
        \caption{mouth\\ShrugUpper}
    \end{subfigure}
    \begin{subfigure}[t]{0.13\linewidth}
        \centering
        \includegraphics[width=\linewidth]{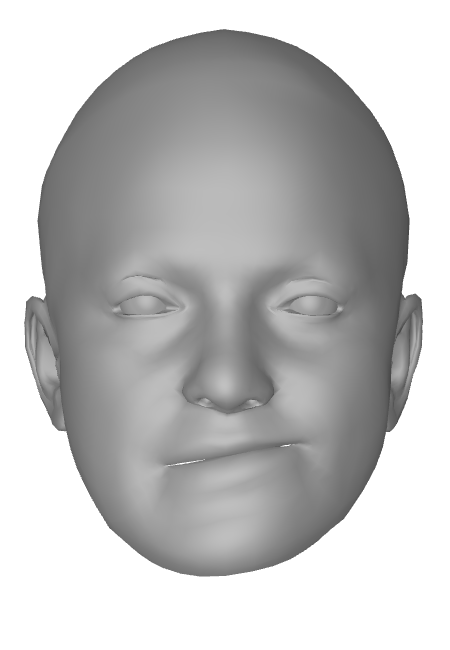}
        \caption{mouth\\PressLeft}
    \end{subfigure}
    \begin{subfigure}[t]{0.13\linewidth}
        \centering
        \includegraphics[width=\linewidth]{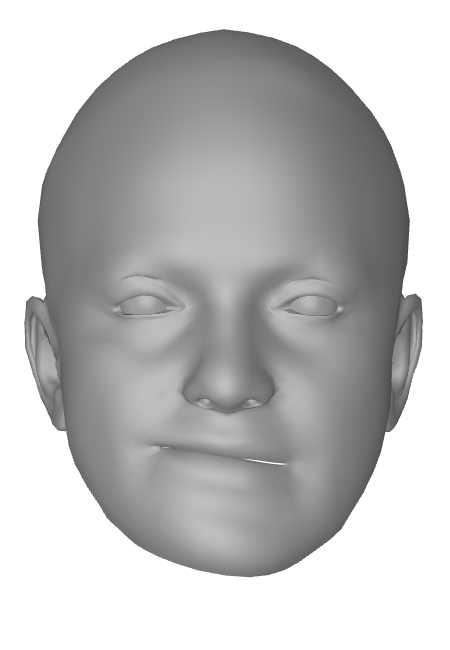}
        \caption{mouth\\PressRight}
    \end{subfigure}
    \begin{subfigure}[t]{0.13\linewidth}
        \centering
        \includegraphics[width=\linewidth]{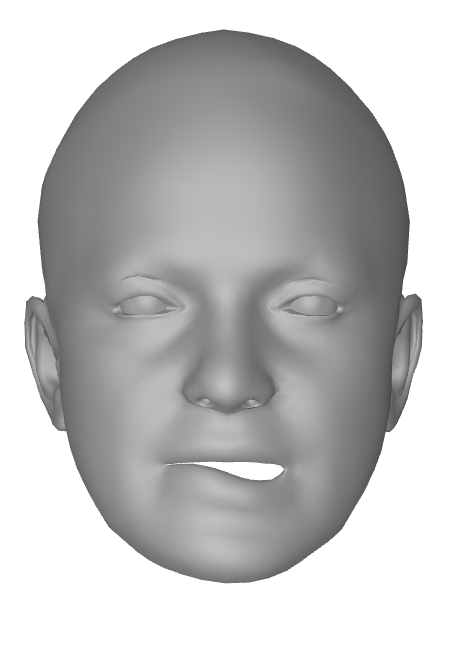}
        \caption{mouthLower\\DownLeft}
    \end{subfigure}
    \begin{subfigure}[t]{0.13\linewidth}
        \centering
        \includegraphics[width=\linewidth]{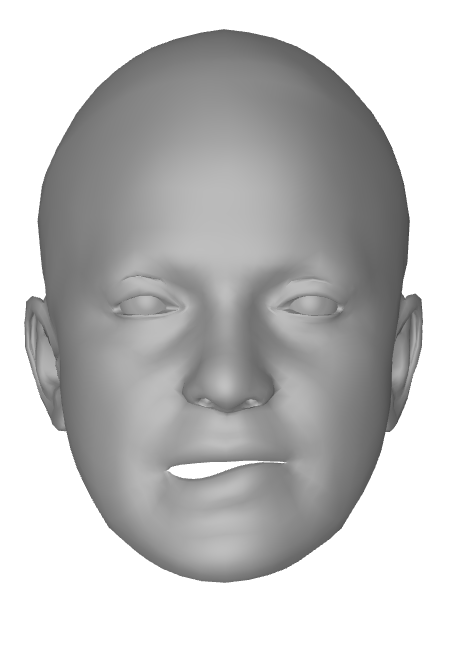}
        \caption{mouthLower\\DownRight}
    \end{subfigure}
    \begin{subfigure}[t]{0.13\linewidth}
        \centering
        \includegraphics[width=\linewidth]{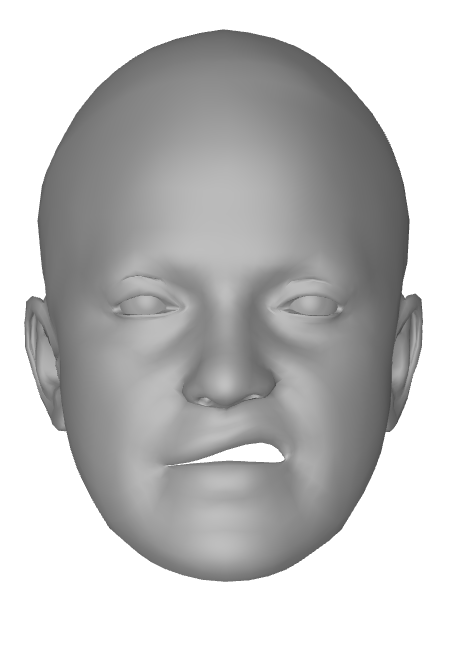}
        \caption{mouthUpper\\UpLeft}
    \end{subfigure}
    \begin{subfigure}[t]{0.13\linewidth}
        \centering
        \includegraphics[width=\linewidth]{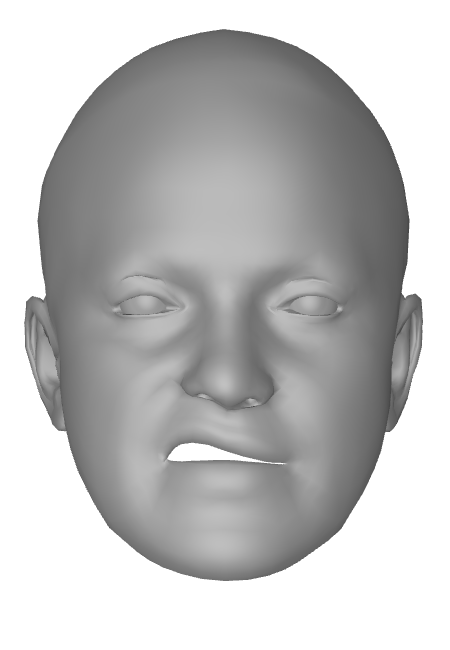}
        \caption{mouthUpper\\UpRight}
    \end{subfigure}
    \begin{subfigure}[t]{0.13\linewidth}
        \centering
        \includegraphics[width=\linewidth]{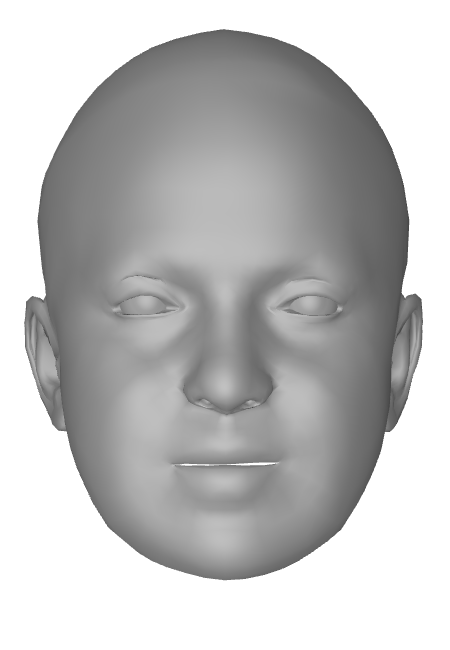}
        \caption{cheekPuff}
    \end{subfigure}
    \begin{subfigure}[t]{0.13\linewidth}
        \centering
        \includegraphics[width=\linewidth]{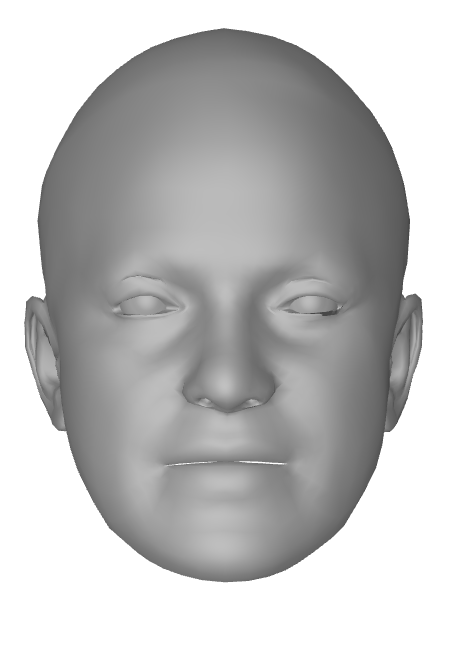}
        \caption{cheek\\SquintLeft}
    \end{subfigure}
    \begin{subfigure}[t]{0.13\linewidth}
        \centering
        \includegraphics[width=\linewidth]{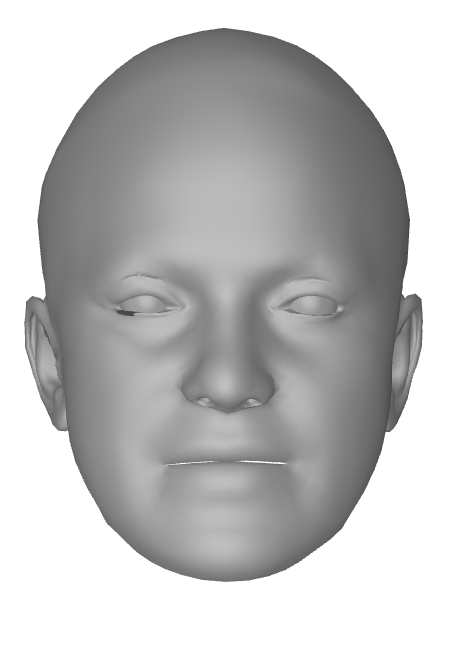}
        \caption{cheek\\SquintRight}
    \end{subfigure}
    \begin{subfigure}[t]{0.13\linewidth}
        \centering
        \includegraphics[width=\linewidth]{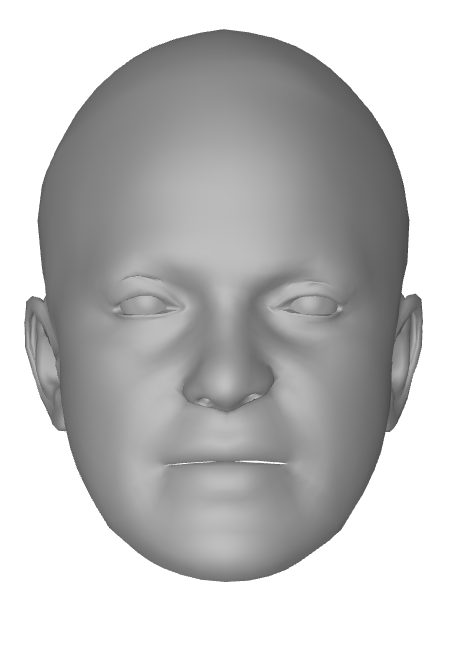}
        \caption{nose\\SneerLeft}
    \end{subfigure}
    \begin{subfigure}[t]{0.13\linewidth}
        \centering
        \includegraphics[width=\linewidth]{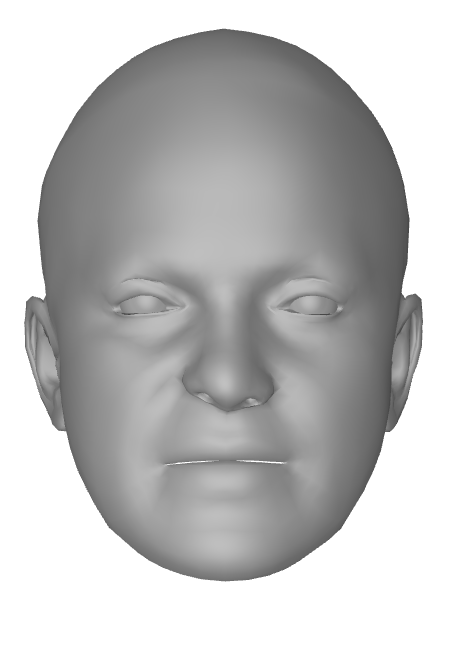}
        \caption{nose\\SneerRight}
    \end{subfigure}
    \caption{
        \textbf{Constructed blendshapes} of the speaker `FaceTalk\_170731\_00024\_TA'.
    }\label{fig:app:blendshape}
\end{figure*}

\begin{figure*}[t]
    \addtocounter{figure}{1}
    \centering
    \includegraphics[width=0.329\linewidth]{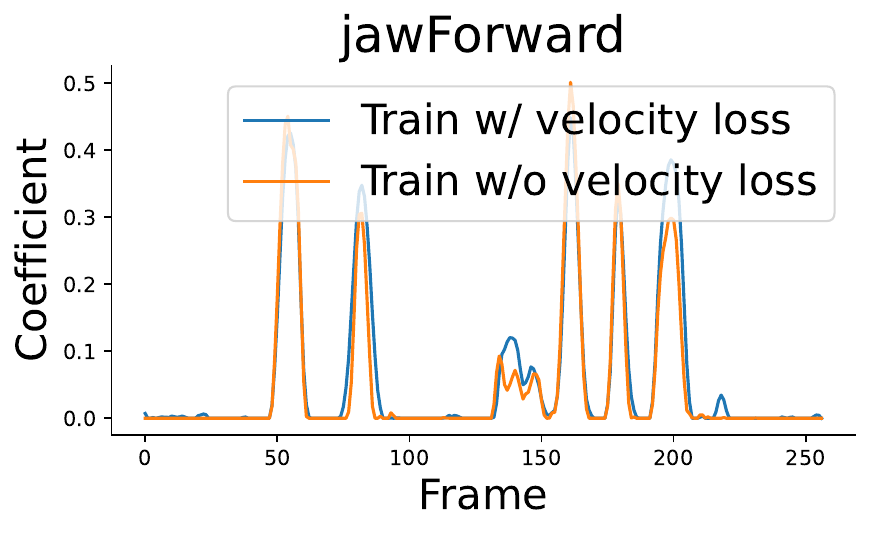}
    \includegraphics[width=0.329\linewidth]{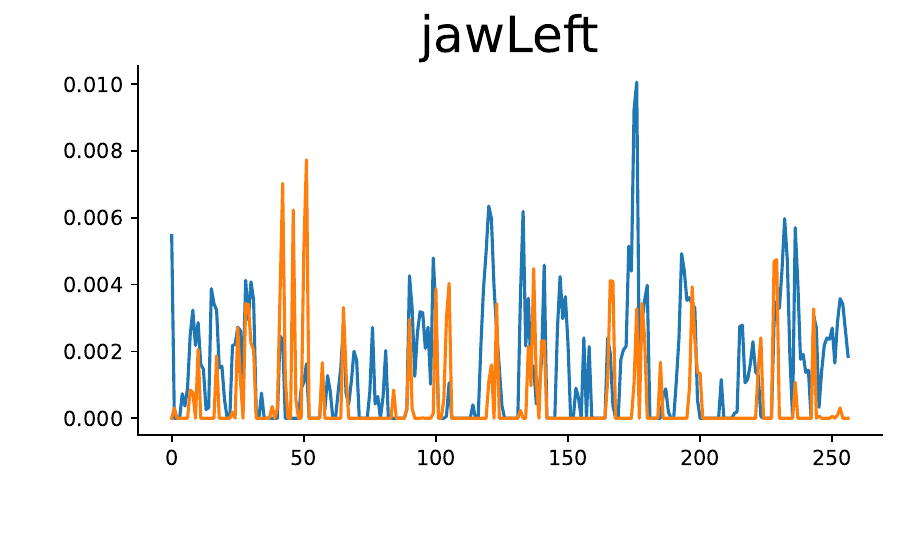}
    \includegraphics[width=0.329\linewidth]{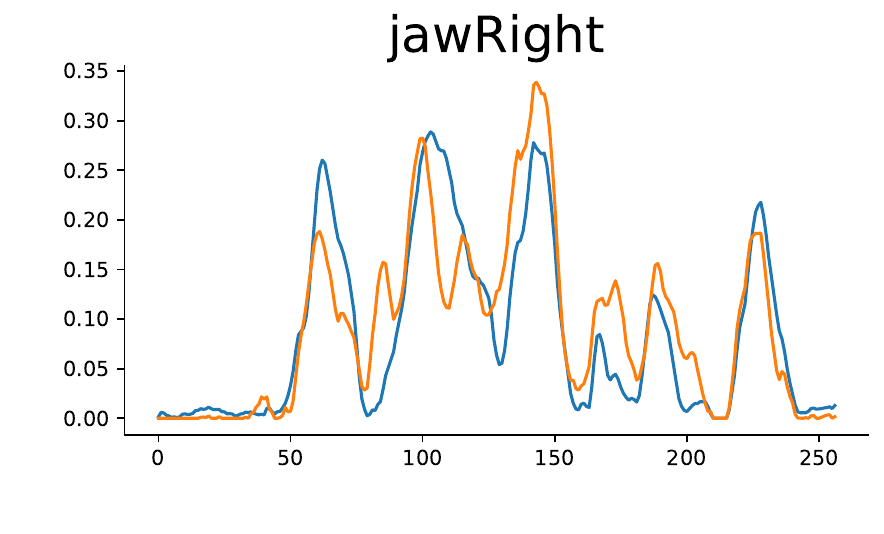}
    \includegraphics[width=0.329\linewidth]{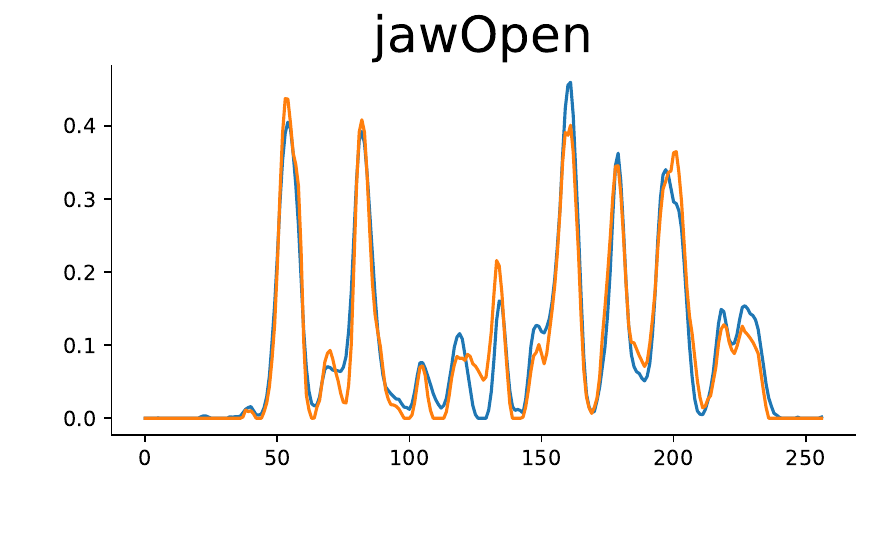}
    \includegraphics[width=0.329\linewidth]{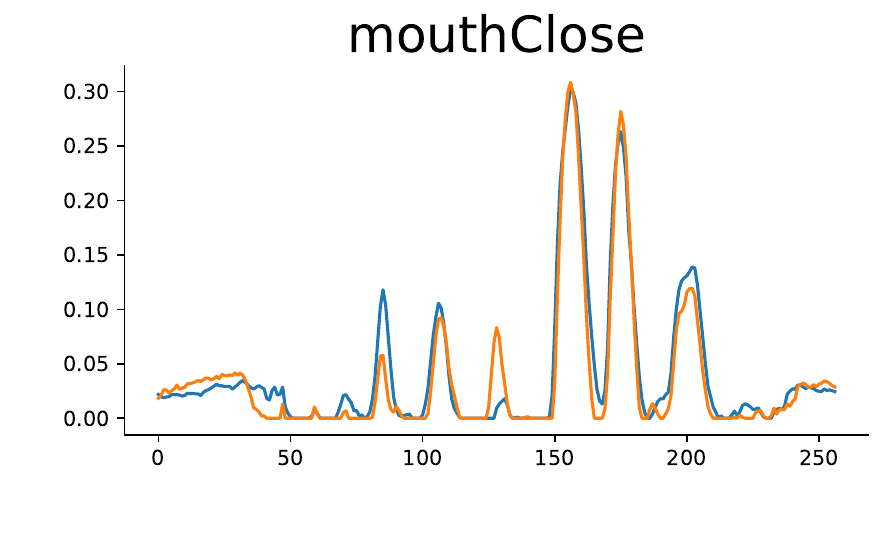}
    \includegraphics[width=0.329\linewidth]{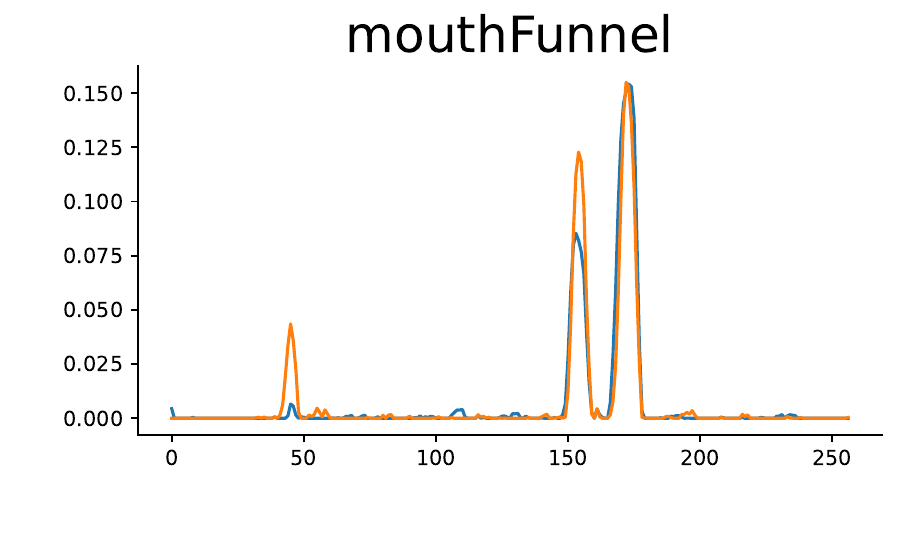}
    \includegraphics[width=0.329\linewidth]{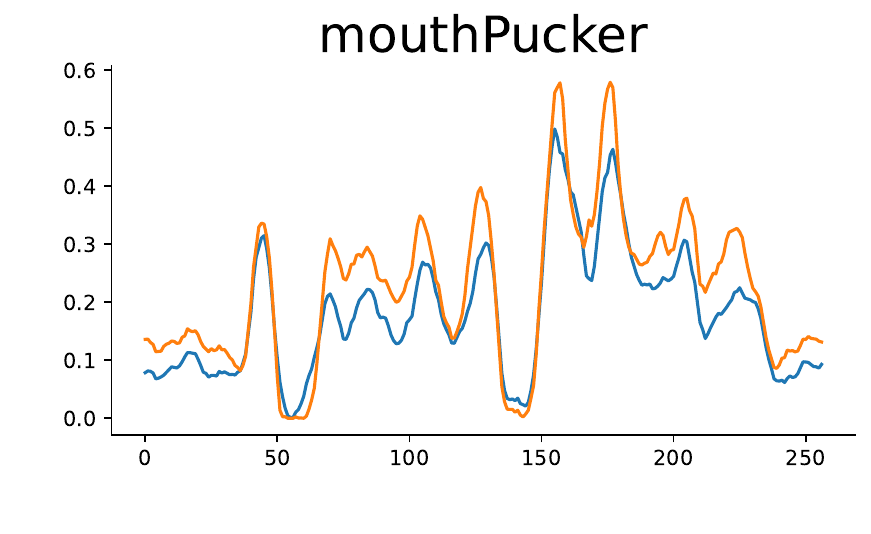}
    \includegraphics[width=0.329\linewidth]{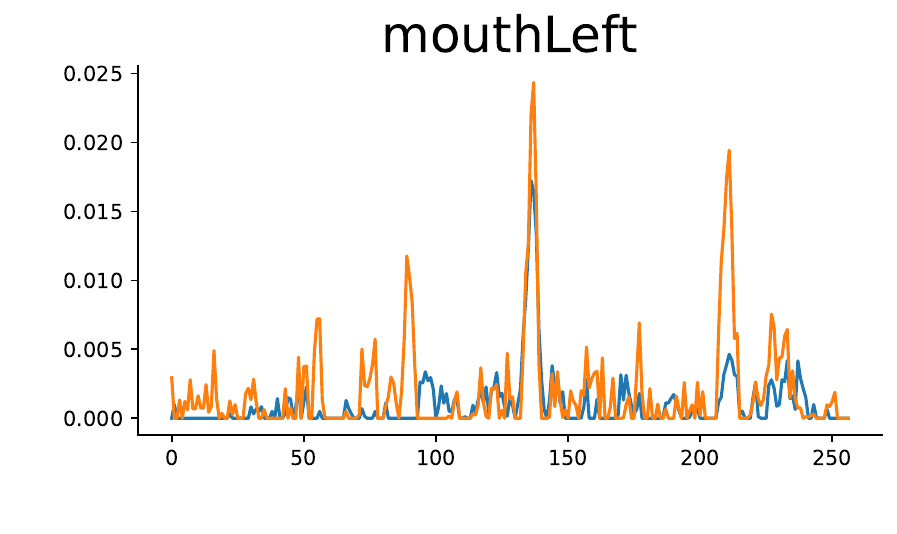}
    \includegraphics[width=0.329\linewidth]{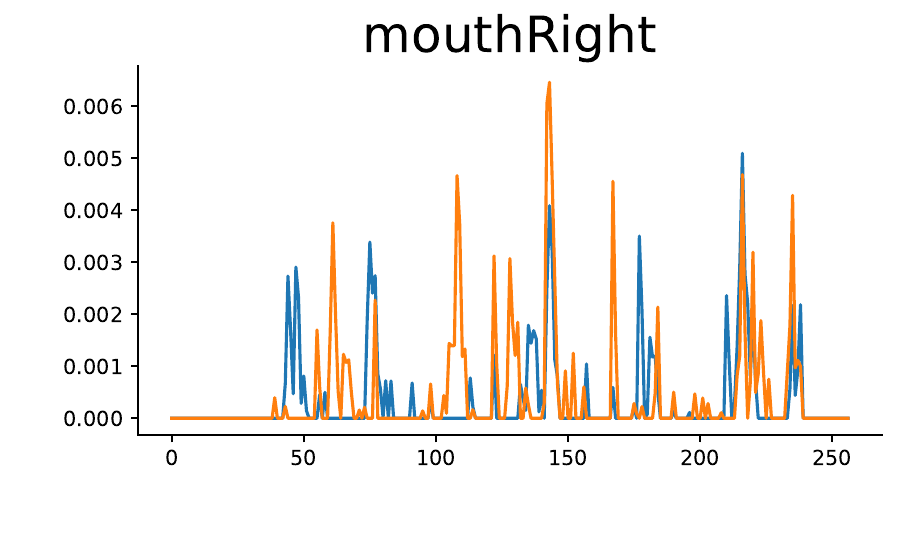}
    \includegraphics[width=0.329\linewidth]{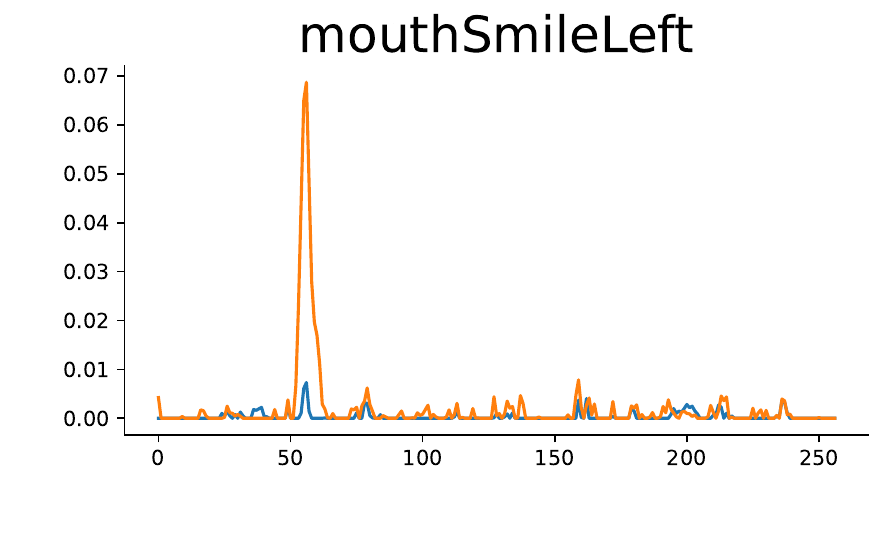}
    \includegraphics[width=0.329\linewidth]{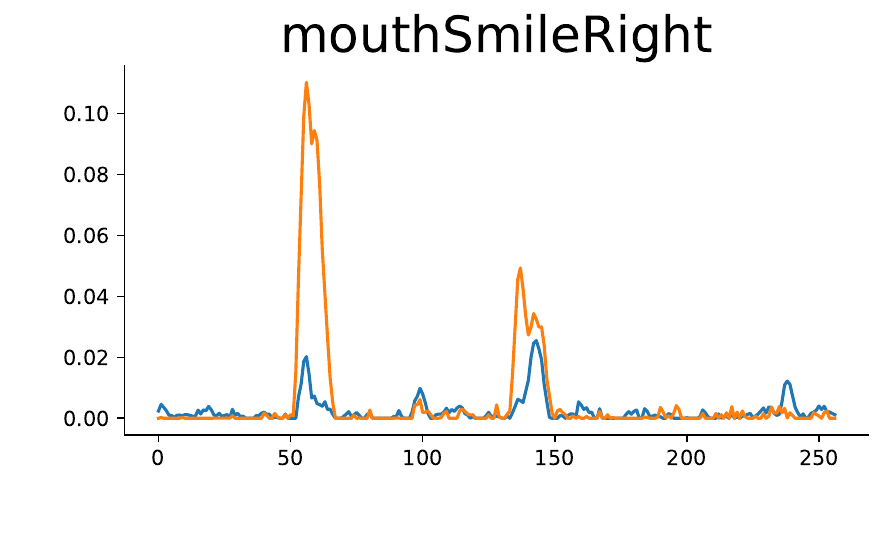}
    \includegraphics[width=0.329\linewidth]{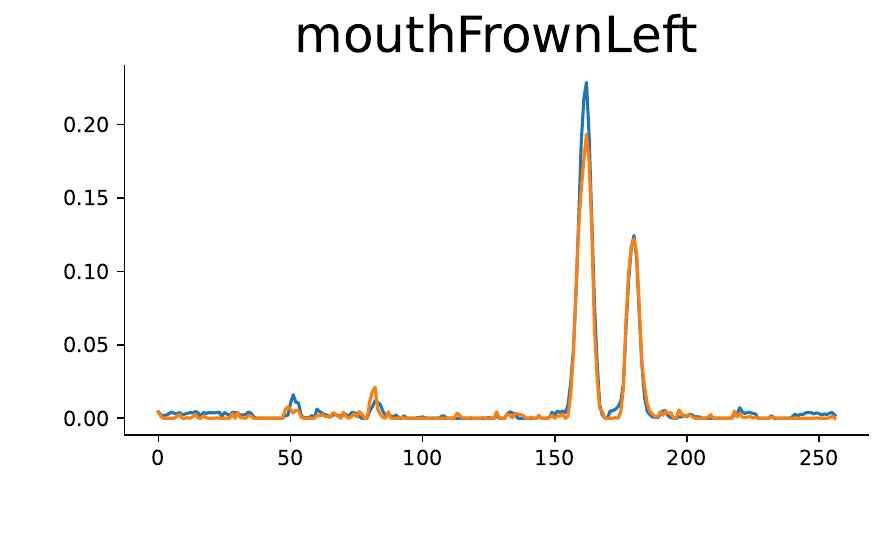}
    \includegraphics[width=0.329\linewidth]{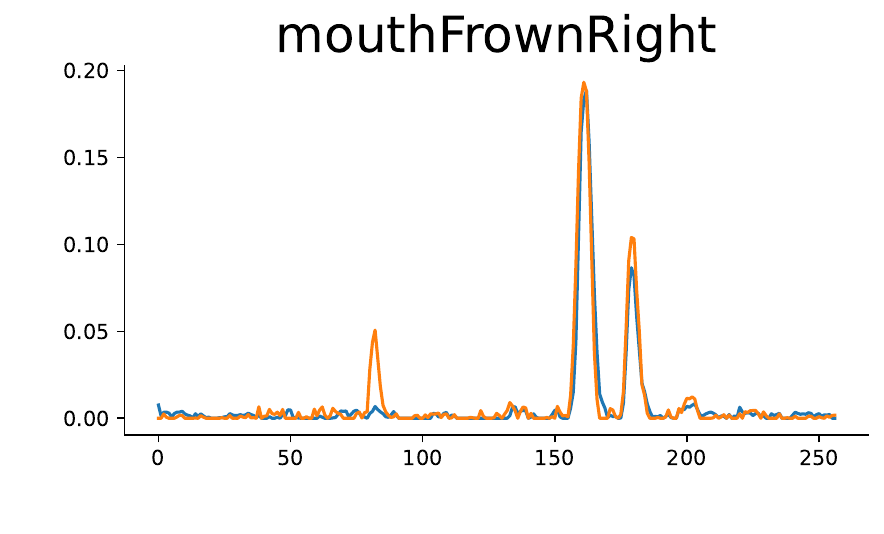}
    \includegraphics[width=0.329\linewidth]{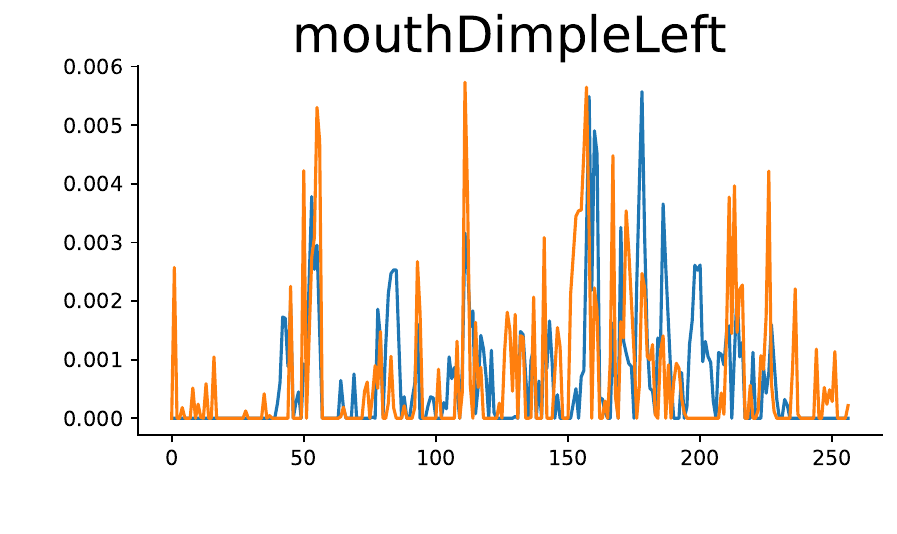}
    \includegraphics[width=0.329\linewidth]{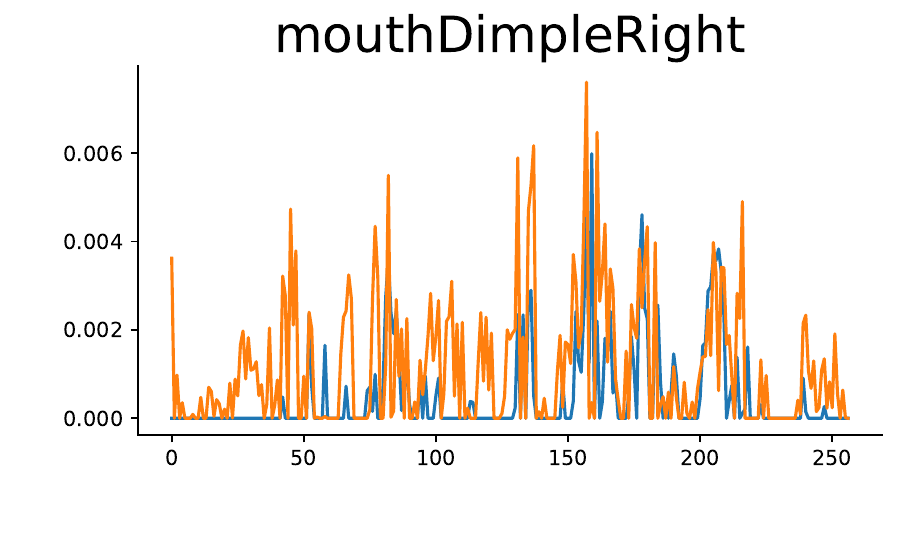}
    \includegraphics[width=0.329\linewidth]{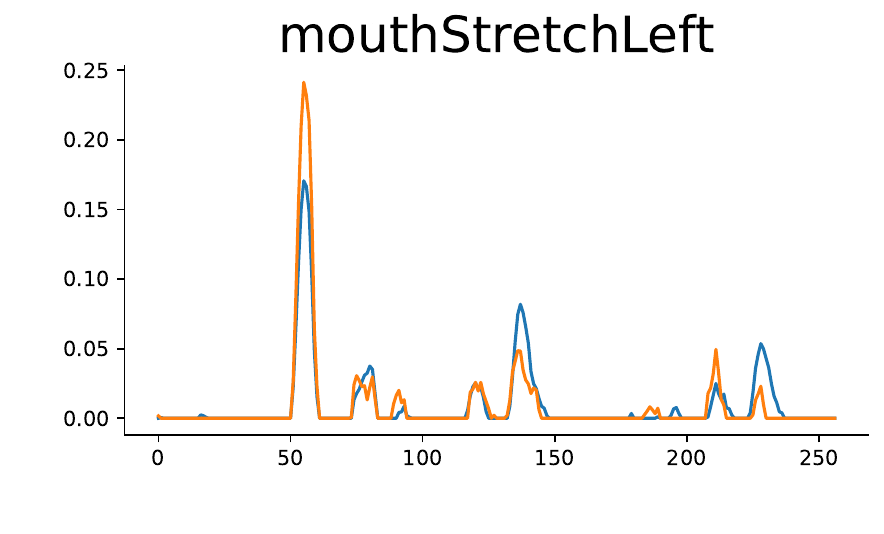}
    \includegraphics[width=0.329\linewidth]{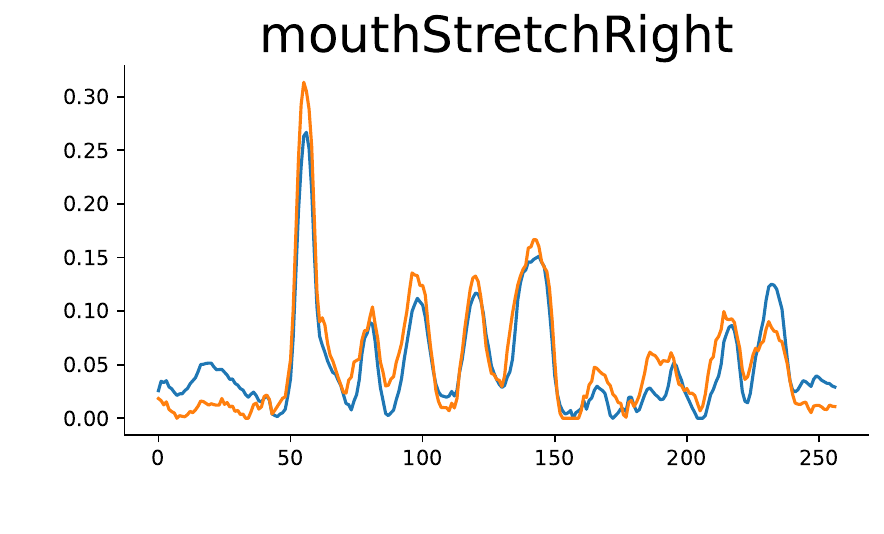}
    \includegraphics[width=0.329\linewidth]{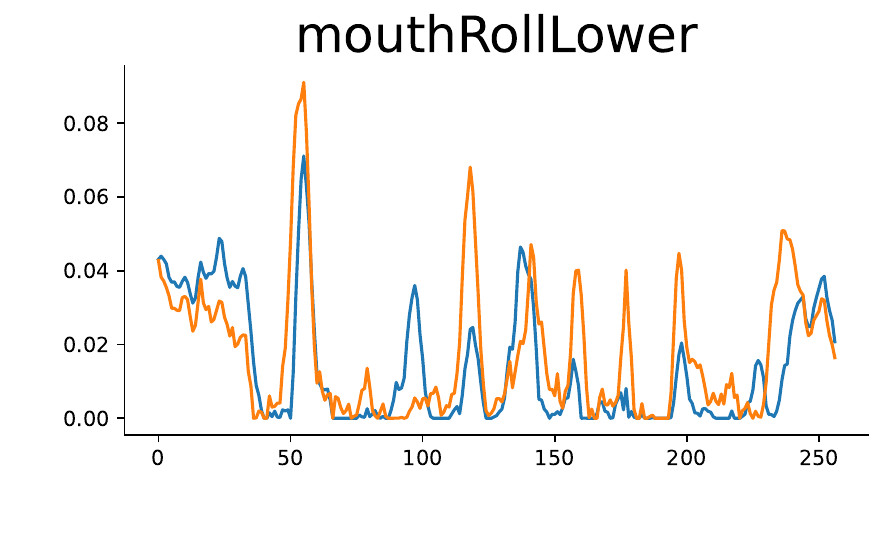}
\end{figure*}
\begin{figure*}[t]\ContinuedFloat
    \centering
    \includegraphics[width=0.329\linewidth]{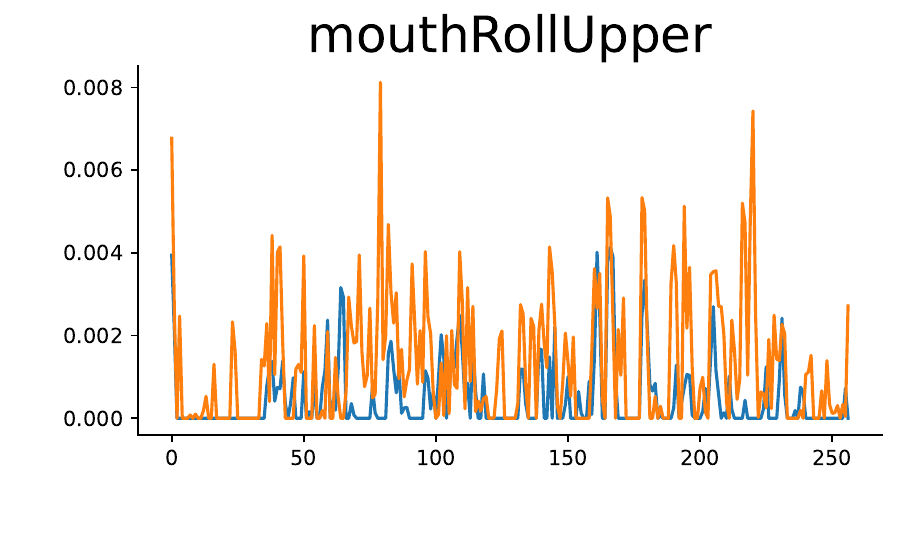}
    \includegraphics[width=0.329\linewidth]{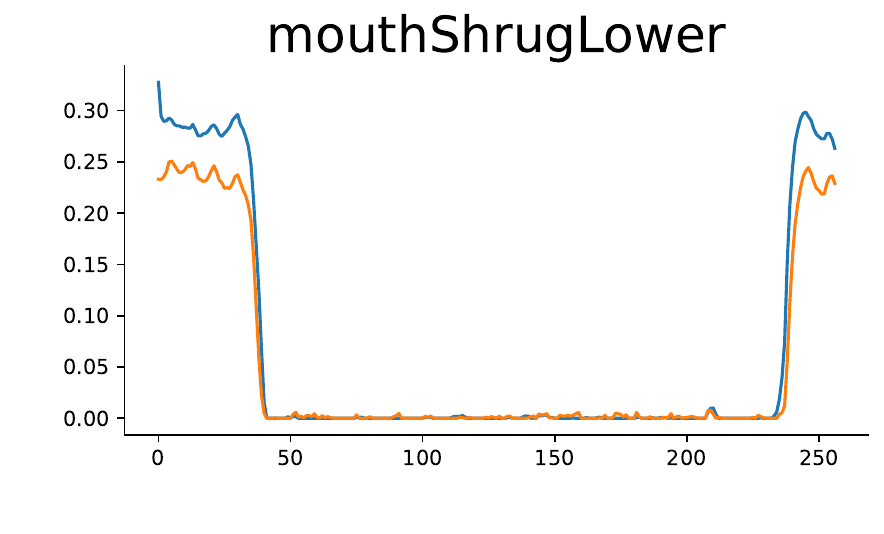}
    \includegraphics[width=0.329\linewidth]{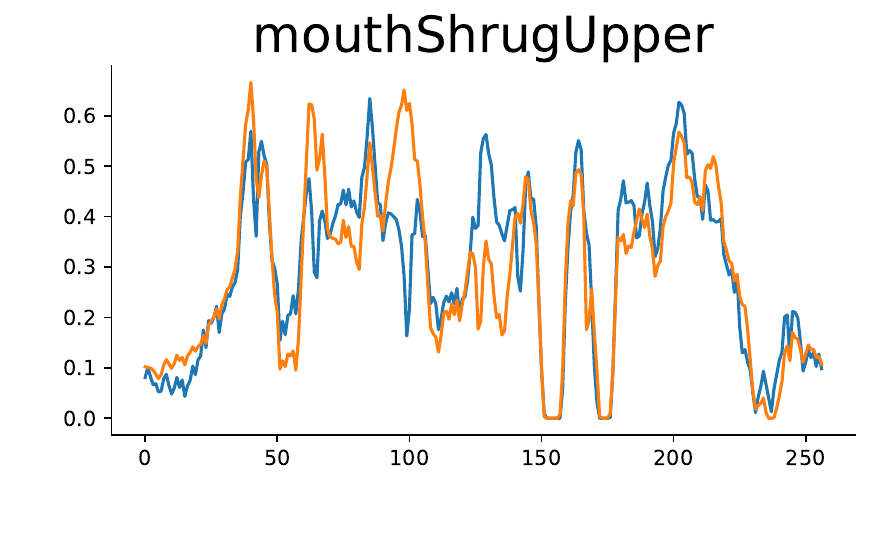}
    \includegraphics[width=0.329\linewidth]{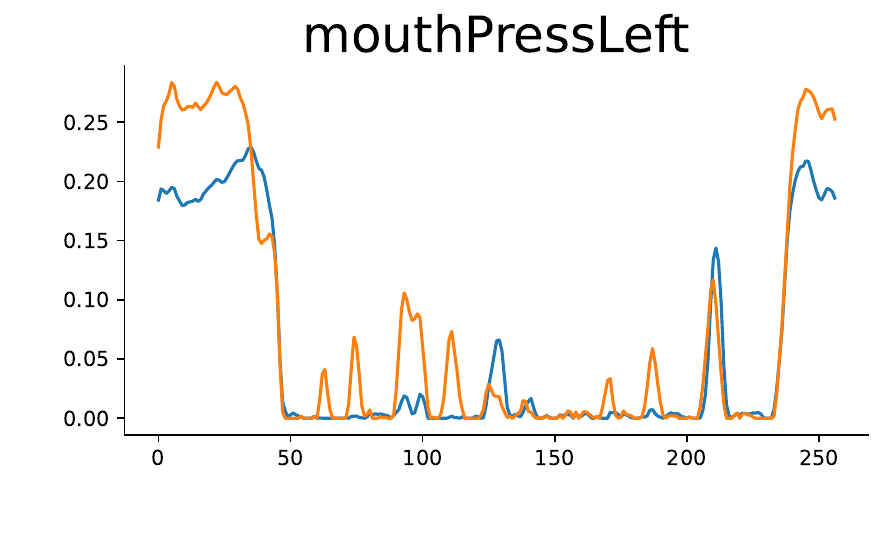}
    \includegraphics[width=0.329\linewidth]{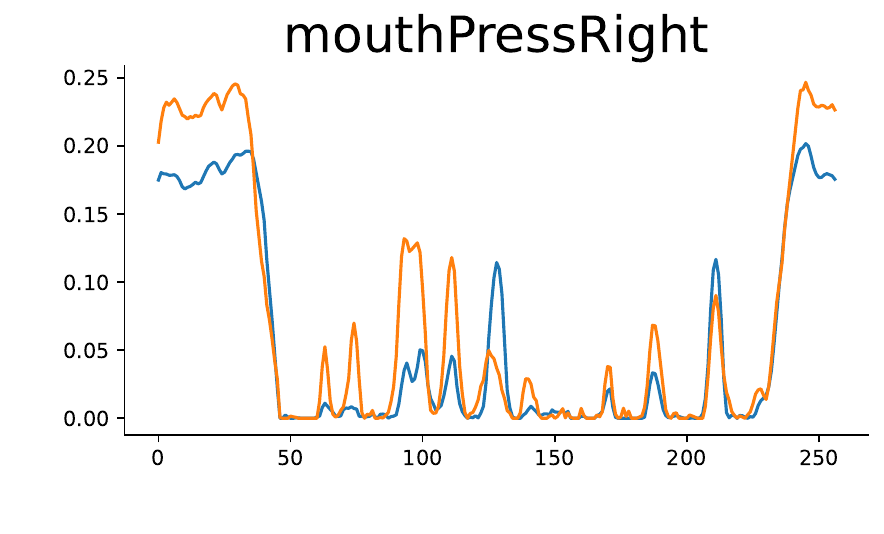}
    \includegraphics[width=0.329\linewidth]{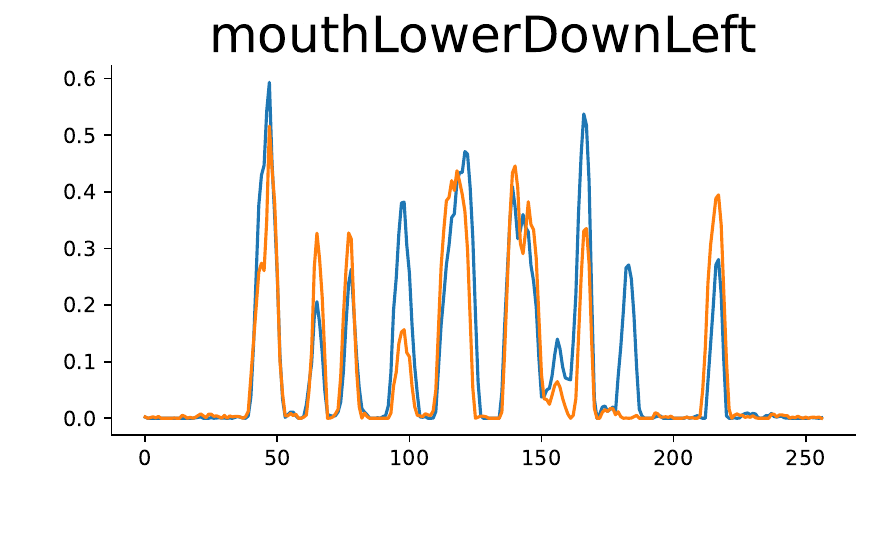}
    \includegraphics[width=0.329\linewidth]{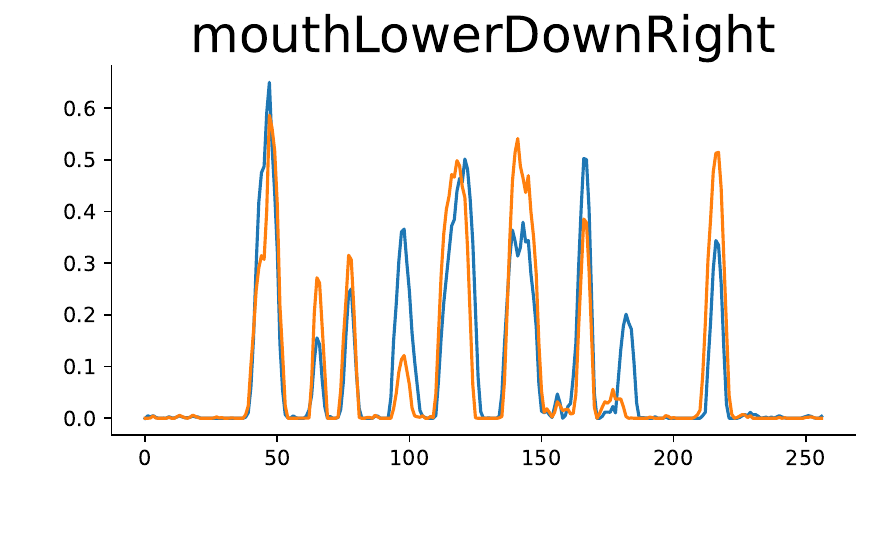}
    \includegraphics[width=0.329\linewidth]{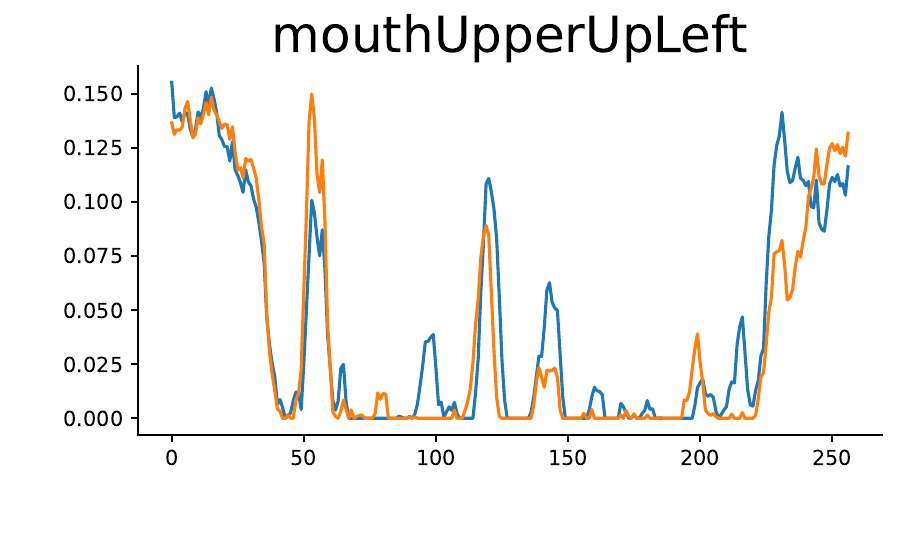}
    \includegraphics[width=0.329\linewidth]{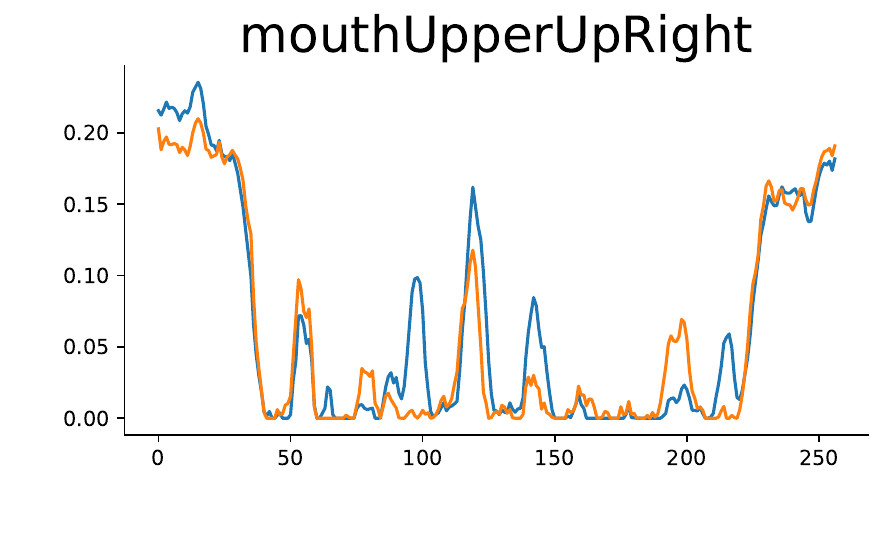}
    \includegraphics[width=0.329\linewidth]{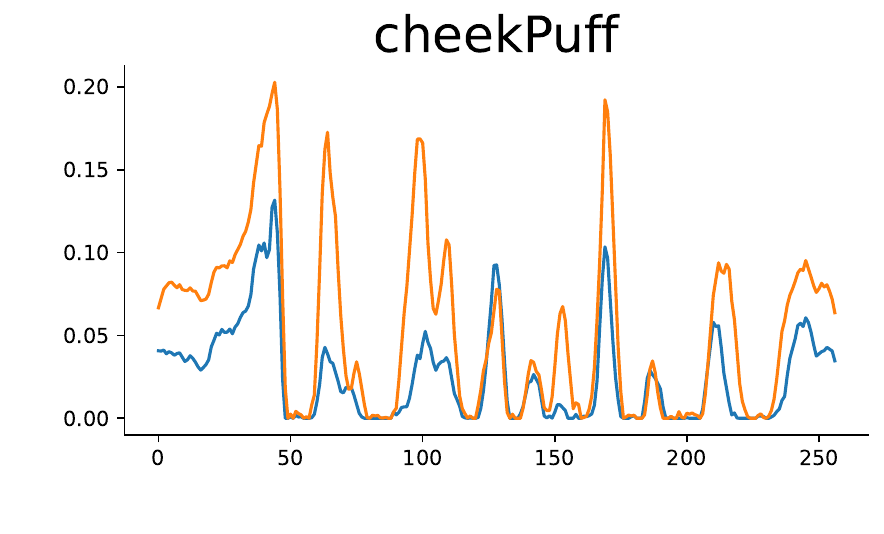}
    \includegraphics[width=0.329\linewidth]{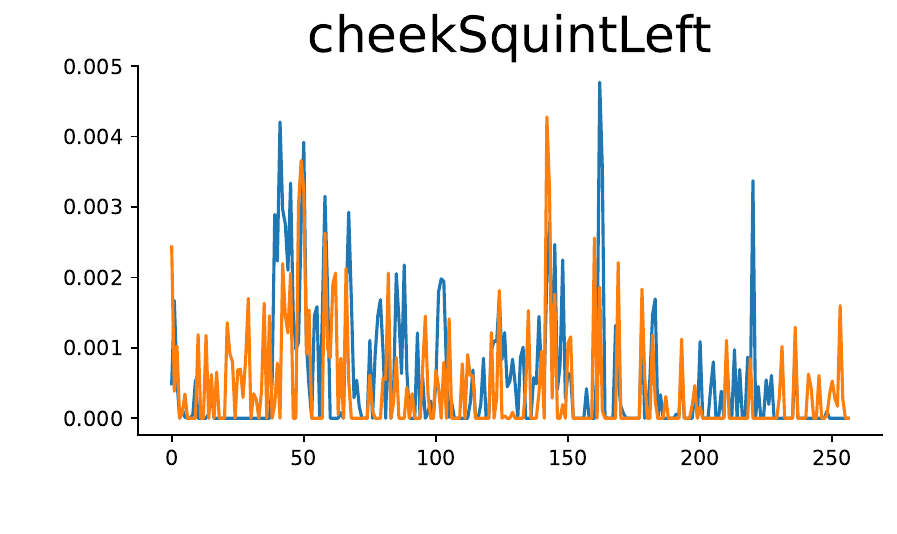}
    \includegraphics[width=0.329\linewidth]{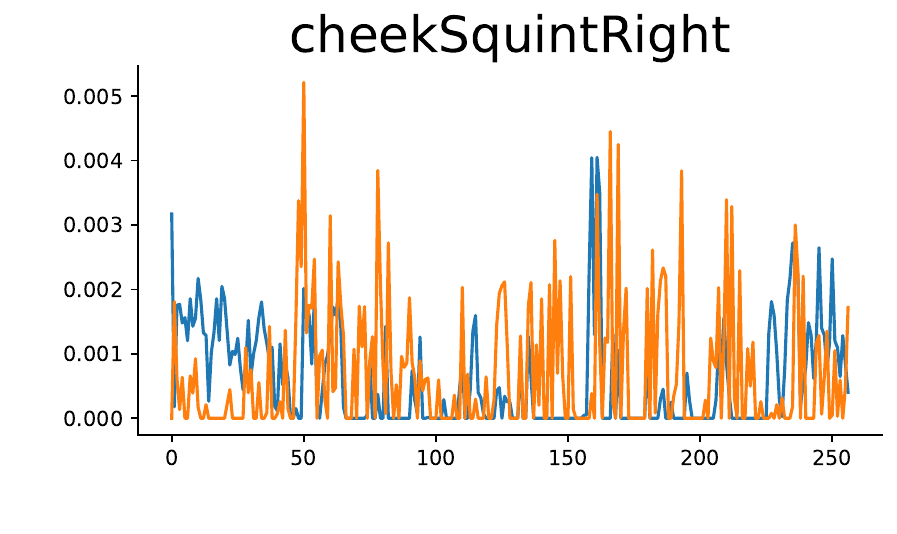}
    \includegraphics[width=0.329\linewidth]{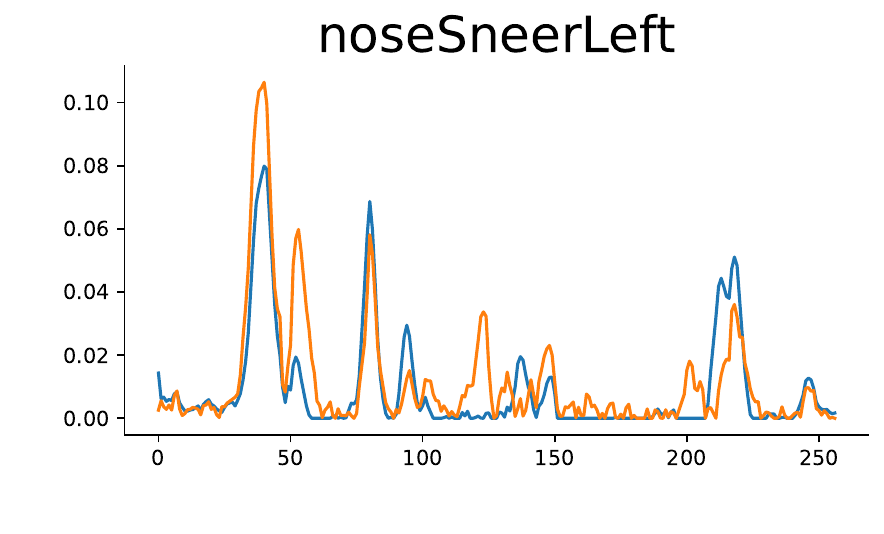}
    \includegraphics[width=0.329\linewidth]{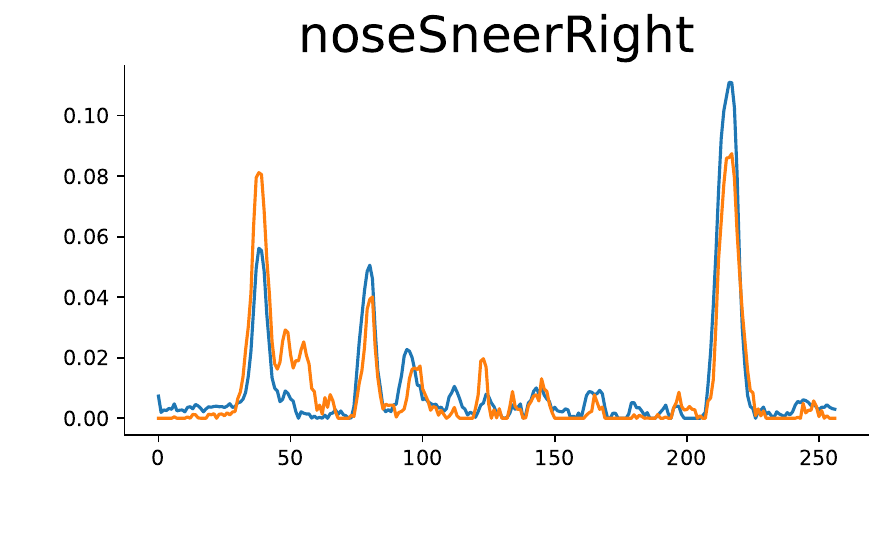}
    \caption{
        \textbf{Effect of the velocity loss - Full version of Fig.~\ref{fig:tc:velocity}}.
        Inference results of SAiD trained with and without velocity loss.
        Each blendshape coefficient sequence is generated using the same random seed, conditioned on the `FaceTalk\_170731\_00024\_TA/sentence01.wav'.
    }\label{fig:app:ablation_velocity}
\end{figure*}

\end{document}